\pdfoutput=1
\documentclass{bmvc2k}
\usepackage{amsfonts}
\usepackage{booktabs}
\usepackage{tabularx,booktabs}
\usepackage{caption}
\usepackage{array}
\usepackage{graphicx}
\usepackage{nicefrac}
\usepackage{xspace}

\usepackage{pgfplots}
\usepackage{pgfplotstable}
\usepackage{pgf,tikz}

\usepackage{scrextend}
\usepackage{hyperref}
\usepackage{bigfoot}
\usepackage{url}

\newcolumntype{L}{>{$}l<{$}}
\newcolumntype{C}{>{$}c<{$}}
\newcolumntype{R}{>{$}r<{$}}

\title{Multiple-Kernel Local-Patch Descriptor}

\addauthor{Arun Mukundan}{arun.mukundan@cmp.felk.cvut.cz}{1}
\addauthor{Giorgos Tolias}{giorgos.tolias@cmp.felk.cvut.cz}{1}
\addauthor{Ond\v{r}ej Chum}{chum@cmp.felk.cvut.cz}{1}
\addinstitution{Visual Recognition Group,\\
                Faculty of Electrical Engineering,\\
                Czech Technical University in Prague}

\runninghead{Mukundan, Tolias, Chum}{Multiple-Kernel Local-Patch Descriptor}

\begin{document}

\newgeometry{twoside,headsep=3mm,papersize={410pt,620pt},inner=15mm,outer=6mm,top=3mm,includehead,bottom=1mm,heightrounded}

\maketitle

\def\real{\mathbb{R}}
\def\integer{\mathbb{Z}}
\def\l2{\ensuremath{\ell_2}\xspace}
\def\linf{\ensuremath{\ell_\infty}\xspace}

\def\des{KD\xspace}

\def\P{\ensuremath{\mathcal{P}}\xspace}
\def\Q{\ensuremath{\mathcal{Q}}\xspace}

\def\p{\ensuremath{p}\xspace}
\def\q{\ensuremath{q}\xspace}

\def\pm{\ensuremath{\p_m}\xspace}
\def\px{\ensuremath{\p_x}\xspace}
\def\py{\ensuremath{\p_y}\xspace}
\def\pt{\ensuremath{\p_\theta}\xspace}
\def\ptt{\ensuremath{\p_{\tilde{\theta}}}\xspace}
\def\pp{\ensuremath{\p_\phi}\xspace}
\def\pr{\ensuremath{\p_\rho}\xspace}
\def\pg{\ensuremath{\p_g}\xspace}

\def\qm{\ensuremath{\q_m}\xspace}
\def\qx{\ensuremath{\q_x}\xspace}
\def\qy{\ensuremath{\q_y}\xspace}
\def\qt{\ensuremath{\q_\theta}\xspace}
\def\qtt{\ensuremath{\q_{\tilde{\theta}}}\xspace}
\def\qp{\ensuremath{\q_\phi}\xspace}
\def\qr{\ensuremath{\q_\rho}\xspace}
\def\qg{\ensuremath{\q_g}\xspace}

\def\n{\ensuremath{\gamma}\xspace}

\def\k{\ensuremath{k}\xspace}
\def\kx{\ensuremath{k_x}\xspace}
\def\ky{\ensuremath{k_y}\xspace}
\def\kr{\ensuremath{k_\rho}\xspace}
\def\kt{\ensuremath{k_\theta}\xspace}
\def\ktt{\ensuremath{k_{\tilde{\theta}}}\xspace}
\def\kp{\ensuremath{k_\phi}\xspace}
\def\km{\ensuremath{k_m}\xspace}
\def\K{\ensuremath{K}\xspace}
\def\VM{\mathsf{VM}}
\def\V{\mathbf{V}}
\def\M{\mathcal{M}}
\def\Vp{\ensuremath{\mathbf{V_{\P}}}\xspace}
\def\Vq{\ensuremath{\mathbf{V_{\Q}}}\xspace}

\def\map{\ensuremath{\psi}\xspace}
\def\mapm{\ensuremath{\psi_m}\xspace}
\def\mapx{\ensuremath{\psi_x}\xspace}
\def\mapy{\ensuremath{\psi_y}\xspace}
\def\mapr{\ensuremath{\psi_\rho}\xspace}
\def\mapt{\ensuremath{\psi_\theta}\xspace}
\def\maptt{\ensuremath{\psi_{\tilde{\theta}}}\xspace}
\def\mapp{\ensuremath{\psi_\phi}\xspace}
\def\mappa{\ensuremath{\psi_{\prma}}\xspace}
\def\mappb{\ensuremath{\psi_{\prmb}}\xspace}

\def\prma{\ensuremath{\phi\rho\tilde{\theta}}\xspace}
\def\prmb{\ensuremath{xy\theta}\xspace}

\def\fr{\omega}
\def\Fr{\Omega}

\def\o{\ensuremath{\phi}\xspace}
\newcommand{\mydelta}{{\mathchoice{\scriptstyle \Delta}{\scriptstyle \Delta}{\scriptscriptstyle \Delta}{\scriptscriptstyle \Delta}}}
\newcommand{\Dp}{\ensuremath{\mydelta \p}\xspace}
\newcommand{\ak}{\ensuremath{\hat{k}}\xspace}
\newcommand{\ith}{{(i)}}
\newcommand{\bOmega}{\ensuremath{\bar{\Omega}}\xspace}

\def\tsp{\hspace{1pt}}
\def\sssp{\hspace{2pt}}
\def\ssp{\hspace{3pt}}
\def\msp{\hspace{5pt}}
\def\bsp{\hspace{7pt}}

\newcommand{\alert}[1]{{\color{red}{#1}}}
\renewcommand{\paragraph}[1]{\vspace{.0\baselineskip}\noindent{\bf #1}\xspace}
\newcommand{\xcaption}[2][1]{\caption{#2}\vspace{-#1\baselineskip}}
\newcommand{\eq}{eqn.\xspace}

\def\etal{\emph{et al.}\xspace}
\def\ie{\emph{i.e.}\xspace}
\def\eg{\emph{e.g.}\xspace}
\def\wrt{\emph{w.r.t.}\xspace}

\def\eg{\emph{e.g}\bmvaOneDot}
\def\Eg{\emph{E.g}\bmvaOneDot}
\def\etal{\emph{et al}\bmvaOneDot}
\def\etc{\emph{etc}\xspace}

\def\xya{{\it cartes}\xspace}
\def\baseline{{\it polar}\xspace}

\def\half{\ensuremath{\nicefrac{1}{2}}\xspace}

\newcommand{\leg}[1]{\addlegendentry{#1}}
\vspace{-21pt}
\begin{abstract}
\vspace{-3pt}
We propose a multiple-kernel local-patch descriptor based on efficient match kernels of patch gradients. It combines two parametrizations of gradient position and direction, each parametrization provides robustness to a different type of patch miss-registration: polar parametrization for noise in the patch dominant orientation detection, Cartesian for imprecise location of the feature point. Even though handcrafted, the proposed method consistently outperforms the state-of-the-art methods on two local patch benchmarks.
\vspace{-12pt}
\end{abstract}


\section{Introduction}
\label{sec:intro}
Representing and matching local features is an essential step of several computer vision tasks.
It has attracted a lot of attention in the last decades, when local features still were a required step of most approaches.
Despite the large focus on Convolutional Neural Networks (CNN) to process whole images, local features still remain important and necessary for tasks such as Structure-from-Motion (SfM)~\cite{FGGJR10}, stereo matching~\cite{MMPL15}, or retrieval under severe change in viewpoint or scale~\cite{SRCF15}.

Recently, the focus has shifted from hand-crafted descriptors to CNN-based descriptors. Learning such descriptors relies on large training sets of patches, that are commonly provided as a side-product of SfM~\cite{WB07}.
Remarkable performance is achieved on a standard benchmark~\cite{BRPM16}.
However, recent work~\cite{BLVM17,SLJS17} shows that CNN-based approaches do not necessarily generalize equally well on different tasks or different datasets.
Hand-crafted descriptors still appear an attractive alternative.

We build upon the hand-crafted kernel descriptor proposed by Bursuc \etal~\cite{BTJ15} that is shown to have good performance, even compared to learned alternatives.
Its few parameters are easily tuned on some validation set, while it is shown to perform well on multiple tasks, as we confirm in our experiments.
Post-processing with PCA and power-law normalization are shown beneficial.

Visualizing and analyzing the parametrization of this kernel descriptor allows us to understand its advantages and disadvantages, mainly the undesirable discontinuity around the patch center.
We propose to combine multiple parametrizations and kernels to achieve robustness to different types of patch miss-registration.
Experimental evaluation shows that the proposed descriptor outperforms all other approaches on two benchmarks designed to compare local-feature descriptors, specifically on the newly introduced HPatches dataset~\cite{BLVM17}, and on the Phototourism benchmark~\cite{WB07}.

\restoregeometry

\section{Related work}
\label{sec:related}
We review prior work on local descriptors, covering both hand-crafted and learned ones.

\textbf{Hand-crafted descriptors} attracted a lot of attention for a decade and a variety of approaches and methodologies exists.
A popular direction is that of gradient histogram-based descriptors, where the most popular representative is SIFT~\cite{L04}. Different variants focus on pooling regions~\cite{LSP05,MS05}, efficiency~\cite{TLF10,AY11}, invariance~\cite{LSP05} or other aspects~\cite{KO08}.
Other are based on filter-bank responses~\cite{KY08}, patch intensity~\cite{CLSF10,RRKB11} or ordered intensity~\cite{OPM02}.

Kernel descriptors based on the idea of Efficient Match Kernels (EMK)~\cite{BS09} encode entities inside a patch (such a gradient, color, \etc) in a continuous domain, rather than as a histogram. The kernels and their few parameters are often hand-picked and tuned on a validation set. Kernel descriptors are commonly represented by a finite-dimensional explicit feature maps. Quantized descriptors, such as SIFT, can be also interpreted as kernel descriptors~\cite{BTJ15,BRF10}.

\textbf{Learned descriptors} commonly require annotation at patch level. Therefore, research in this direction is facilitated by the release of datasets that are originate from an SfM system~\cite{WB07,PDHM+15}. Such training datasets allow effective learning of local descriptors, and in particular, their pooling regions~\cite{WB07,SVZ13}, filter banks~\cite{WB07}, transformations for dimensionality reduction~\cite{SVZ13} or embeddings~\cite{PISZ10}.

Kernelized descriptors are formulated within a supervised framework by Wang \etal~\cite{WWZX+13}, where image labels enable kernel learning and dimensionality reduction.
In this work, we rather focus on minimal learning in the form of discriminatively learned projections. This is several orders of magnitude faster to learn than other learning approaches.

Recently, learning local descriptor is dominated by deep learning. The network architectures are smaller than the corresponding ones performing on images, and use a large amount of training patches. Among representative examples is the work of Simo-Serra \etal~\cite{STFK+15} training with hard positive and negative examples or the work of Zagoruyko~\cite{ZK15} where a central-surround representation is found to be immensely beneficial. CNN-based approaches are seen as joint feature, filter bank, and metric learning~\cite{HLJS+15}. Finally, the state of the art consists of shallower architectures with improved ranking loss~\cite{BRPM16,BJTM16}.
Despite obtaining impressive results on a standard benchmark, CNN-based approaches do not generalize well to other datasets and tasks~\cite{BLVM17,SLJS17}.

A post-processing step is common to both hand-crafted and learned descriptors. This post-processing ranges from simple \l2 normalization, PCA dimensionality reduction, to transformations learned on annotated data.

\vspace{-10pt}
\section{Preliminaries}
\label{sec:prem}
\textbf{Kernelized descriptors.}
In general lines we follow the formulation of Bursuc \etal~\cite{BTJ15}.
We represent a patch \P as a set of pixels $\p \in \P$ and compare two patches \P and \Q via match kernel
\begin{equation}
\M(\P,\Q) = \sum_{\p\in \P} \sum_{\q\in \Q} \k(\p,\q),
\end{equation}
where kernel $\k: \real^n \times \real^n \rightarrow \real$ is a similarity function, typically non-linear, comparing two pixels.
EMK uses an explicit feature map $\map: \real^n \rightarrow \real^d$ to approximate this result as
\begin{equation}
\M(\P,\Q) = \sum_{\p\in \P} \sum_{\q\in \Q} \k(\p,\q) \approx \sum_{\p\in \P} \sum_{\q\in \Q} \map(\p)^\top \map(\q) = \sum_{\p\in \P} \map(\p)^\top \sum_{\q\in \Q} \map(\q).
\end{equation}
Vector $\V(\P) = \sum_{\p\in \P} \map(\p)$ is a \emph{kernelized descriptor} (\des), associated with patch \P, used to approximate $\M(\P,\Q)$, whose explicit evaluation is costly. The approximation is given by a dot product $\V(\P)^\top \V(\Q)$, where $\V(\P)\in \real^d$.
To ensure a unit self similarity, \l2 normalization by a factor \n is introduced. The normalized \des is then given by $\bar{\V}(\P) = \gamma(\P) \V(\P)$, where $\gamma(\P) = (\V(\P)^\top \V(\P))^{-\half}$.

Kernel \k comprises product of kernels that act on scalar pixel attributes
\begin{equation}
\k(\p,\q) = \k_1(\p_1,\q_1) \k_2(\p_2,\q_2) \ldots \k_n(\p_n,\q_n),
\end{equation}
where kernel $\k_n$ is pairwise similarity function for scalars and $\p_n$ are pixel attributes such as position and gradient orientation. Feature map $\map_n$ corresponds to kernel $\k_n$ and feature map $\map$ is constructed via Kronecker product of individual feature maps $\map(\p) = \map_1(\p_1) \otimes \map_2(\p_2) \otimes \ldots \otimes \map_n(\p_n)$. Due to the mixed product property it holds that $\map(\p)^\top \map(\q) \approx \k_1(\p_1,\q_1) \k_2(\p_2,\q_2) \ldots \k_n(\p_n,\q_n)$.

\vspace{7pt}
\textbf{Feature maps.}
As non-linear kernel for scalars we use the normalized Von Mises probability density function\footnote{Also known as the periodic normal distribution}, which is used for image~\cite{TBFJ15} and patch~\cite{BTJ15} representation.
It is parametrized by $\kappa$ controlling the shape of the kernel, where lower $\kappa$ corresponds to wider kernel. We use a stationary kernel that, by definition, depends only on the difference $\Delta_{n} = \p_n-\q_n$, \ie $\k_\VM(\p_n, \q_n) := \k_\VM(\Delta_n)$. We adopt a Fourier series approximation with $N$ frequencies that produces a feature map $\map_\VM: \real \rightarrow \real^{2N+1}$. It has the property that $\k_\VM(\p_n, \q_n) \approx \map_\VM(\p_n)^\top \map_\VM(\q_n)$. The reader is encouraged to read prior work for details on these feature maps~\cite{VZ10}, which are previously used in various contexts~\cite{TBFJ15,BTJ15}.

\vspace{7pt}
\textbf{Descriptor post-processing.}
It is known that further descriptor post-processing~\cite{RTC16,BL15,BTJ15} is beneficial. In particular, \des is further centered and projected as
\begin{equation}
\hat{\V}(\P) = A^\top(\bar{\V}(\P) -\mu),
\label{equ:norm}
\end{equation}
where $\mu \in \real^d$ and $A\in \real^{d\times d}$ are the mean vector and the projection matrix.
These are commonly learned by PCA~\cite{JC12} or with supervision~\cite{RTC16}.
The final descriptor is always \l2-normalized in the end.

\section{Method}
\label{sec:method}
In this section we consider different patch parametrizations and kernels that result in different patch similarity. We discuss the benefits of each and propose how to combine them. We further learn descriptor transformation with supervision and provide useful insight on how patch similarity is affected.

\vspace{7pt}
\textbf{Patch attributes.}
We consider a pixel \p to be associated with coordinates \px, \py in Cartesian coordinate system, coordinates \pr, \pp in polar coordinate system, pixel gradient magnitude \pm, and pixel gradient angle \pt. Angles \pt, \pp $\in [0, 2\pi]$, distance from the center \pr is normalized to $[0, 1]$, while coordinates \px, \py $\in \{1, 2, \ldots, W\}$ for $W \times W$ patches. In order to use feature map $\map_\VM$, attributes \pr, \px, and \py are linearly mapped to $[0, \pi]$. The gradient angle is expressed \wrt the patch orientation, \ie \pt directly, or \wrt to the position of the pixel. The latter is given as $\ptt = \pt - \pp$.

\begin{figure}
\begin{tabular}{ccc}

\begin{tikzpicture}
 \tikzstyle{every node}=[font=\normalsize]
   \begin{axis}[%
   	  title={$\kappa = 8, N = 3$},
      height=0.25\textwidth,
      width=0.33\textwidth,
      xlabel={$\Delta\theta$~~or~~$\Delta\tilde{\theta}$},
      ylabel={$\kt$~~or~~$\ktt$},
      legend pos=north east,
      legend cell align=left,
      legend style={font=\tiny},
      ytick = {0, 1},
      xtick={0, 0.7854, 1.5708, 2.3562, 3.14159},
      xmin = 0, xmax = 3.14159,
      xticklabels={$0$,$\frac{\pi}{4}$,$\frac{\pi}{2}$,$\frac{3\pi}{4}$, $\pi$},
      y label style={at={(axis description cs:0.25,0.5)}},
      title style={at={(axis description cs:0.5,.8)}},
   ]
   \addplot[color=red, style=solid, line width=2pt] table[x index=0,y index=1]{fig/data/n3kappa8.dat};
   \end{axis}
\end{tikzpicture} 
\hspace{5pt}
\begin{tikzpicture}
 \tikzstyle{every node}=[font=\normalsize]
   \begin{axis}[%
   	  title={$\kappa = 8, N = 2$},
      height=0.25\textwidth,
      width=0.33\textwidth,
      xlabel={$\Delta\phi$~~or~~$\Delta\rho$},
      ylabel={$\kp$~~or~~$\kr$},
      legend pos=north east,
      legend cell align=left,
      legend style={font=\tiny},
      xtick={0, 0.7854, 1.5708, 2.3562, 3.14159},
      ytick = {0, 1},
      xmin = 0, xmax = 3.14159,
      xticklabels={$0$,$\frac{\pi}{4}$,$\frac{\pi}{2}$,$\frac{3\pi}{4}$, $\pi$},
      y label style={at={(axis description cs:0.25,0.5)}},
      title style={at={(axis description cs:0.5,.8)}},      
   ]
   \addplot[color=red, style=solid, line width=2pt] table[x index=0,y index=1]{fig/data/n2kappa8.dat};
   \end{axis}
\end{tikzpicture} 
\hspace{5pt}
\begin{tikzpicture}
 \tikzstyle{every node}=[font=\normalsize]
   \begin{axis}[%
      title={$\kappa = 1, N = 1$},
      height=0.25\textwidth,
      width=0.33\textwidth,
      xlabel={$\Delta x$~~or~~$\Delta y$},
      ylabel={$\kx$~~or~~$\ky$},
      legend pos=north east,
      legend cell align=left,
      legend style={font=\tiny},
      xtick={0, 0.7854, 1.5708, 2.3562, 3.14159},
      ytick = {0, 1},
      xmin = 0, xmax = 3.14159,
      xticklabels={$0$,$\frac{\pi}{4}$,$\frac{\pi}{2}$,$\frac{3\pi}{4}$, $\pi$},
      y label style={at={(axis description cs:0.25,0.5)}},
      title style={at={(axis description cs:0.5,.8)}},    
   ]
   \addplot[color=red, style=solid, line width=2pt] table[x index=0,y index=1]{fig/data/n1kappa1.dat};
   \end{axis}
\end{tikzpicture}

\end{tabular}
\vspace{5pt}
\caption{Kernel approximations that we use for pixel attributes. Parameter $\kappa$ and the number of frequencies $N$ define the final shape. The choice of kernel parameters is guided by~\cite{BTJ15}.   \label{fig:kernels}}
\end{figure}
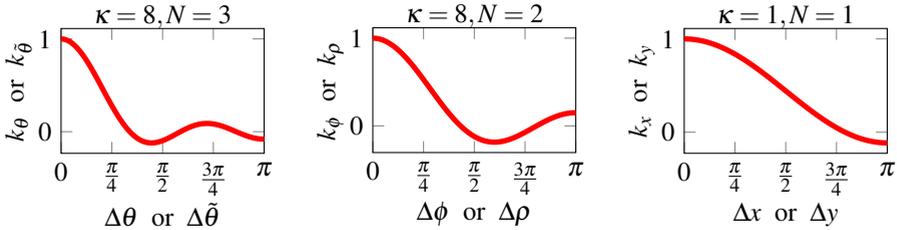
\textbf{Patch parametrizations.}
Composing patch kernel \k as a product of kernels over different attributes enables easy design of various patch similarities. Correspondingly, this defines different \des.
All attributes \px, \py, \pr, \pt, \pp, and \ptt are matched by the Von Mises kernel, namely, \kx, \ky, \kr, \kt, \kp, and \ktt parameterized by $\kappa_x$, $\kappa_y$, $\kappa_\rho$, $\kappa_\theta$, $\kappa_\phi$, and $\kappa_{\tilde{\theta}}$, respectively.

In this work we focus on the two following match kernels over patches. One in \emph{polar} coordinates
\begin{equation}
\M_{\prma}(\P,\Q) = \sum_{\p\in \P} \sum_{\q\in \Q} \pg\qg \sqrt{\pm}\sqrt{\qm}\kp(\pp,\qp)\kr(\pr,\qr)\ktt(\ptt,\qtt),
\end{equation}
and one in \emph{cartesian} coordinates
\begin{equation}
\M_{\prmb}(\P,\Q) = \sum_{\p\in \P} \sum_{\q\in \Q} \pg\qg \sqrt{\pm}\sqrt{\qm}\kx(\px,\qx)\ky(\py,\qy)\kt(\pt,\qt),
\end{equation}
where $\pg = exp(-\pr^2)$ gives more importance to central pixels, in a similar manner to SIFT.

The \des for the two cases are given by
\begin{align}
\V_{\prma}(\P) &= \sum_{\p\in \P} \pg \pm \mapp(\pp)\otimes \mapr(\pr) \otimes \maptt(\ptt) &= \sum_{\p\in \P} \pg\sqrt{\pm} \mappa(\p)\\
\V_{\prmb}(\P) &= \sum_{\p\in \P} \pg \pm \mapx(\px)\otimes \mapy(\py) \otimes \mapt(\pt) &= \sum_{\p\in \P} \pg\sqrt{\pm}\mappb(\p).
\end{align}
The $\V_{\prma}$ variant is exactly the one proposed by Bursuc \etal~\cite{BTJ15}, considered as a baseline in this work.
Different parametrizations result in different patch similarity, which is analyzed in the following.
In Figure~\ref{fig:kernels} we present the approximation of kernels used per attribute.

\textbf{Descriptor post-processing with supervision.}
Mean vector $\mu$ and projection matrix $A$ can be learned in an unsupervised way, \eg by PCA on a sample descriptor set.
In such case, matrix $A$ is formed by the eigenvectors as columns.
This is the case in prior work, not only for local descriptors~\cite{BTJ15} but also for global image representation~\cite{JC12}.
It was previously observed, and our experiments confirm, that discriminative projection~\cite{MM07} learned on labeled data outperforms post-processing by generative model, such as PCA.
The discriminative projection is composed of two parts, a whitening part and a rotation part.
The whitening part is obtained from the intraclass (matching pairs) covariance matrix, while the rotation part is the PCA of the interclass (non-matching pairs) covariance matrix in the whitened space.
Vector $\mu$ is the mean descriptor vector. To reduce the descriptor dimensionality, only eigenvectors corresponding to the largest eigenvalues are used. We refer to this transformation as learned (supervised) whitening (LW) in the rest of the paper.

\textbf{Visualization of patch similarity.}
We define pixel similarity $\M(\p,\q)$ as kernel response between pixels \p and \q, approximated as $\M(\p,\q) \approx \map(\p)^\top \map(\q)$. To show a spatial distribution of the influence of pixel \p, we define a \emph{patch map} of pixel \p. The patch map has the same size as the image patches, for each pixel \q of the patch, map $\M(\p,\q)$ is evaluated for some constant value of $\q_\theta$.

\begin{figure*}
\centering
\begin{tabular}{@{\ssp}c@{\ssp}ccccc}
   \rotatebox{90}{$\hspace{15pt}\Delta\theta = 0$} &
   \includegraphics[height=75pt]{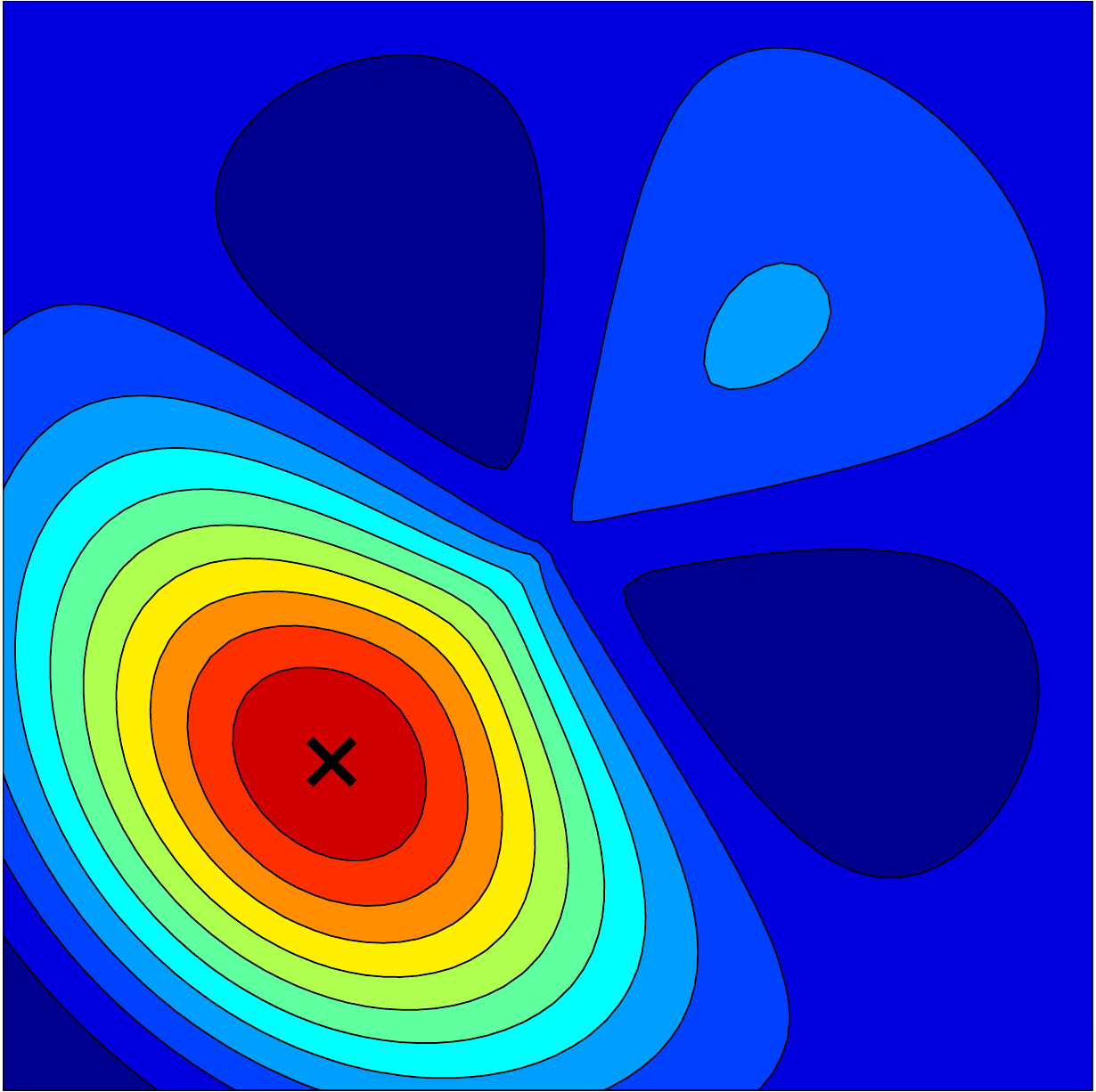}&
   \includegraphics[height=75pt]{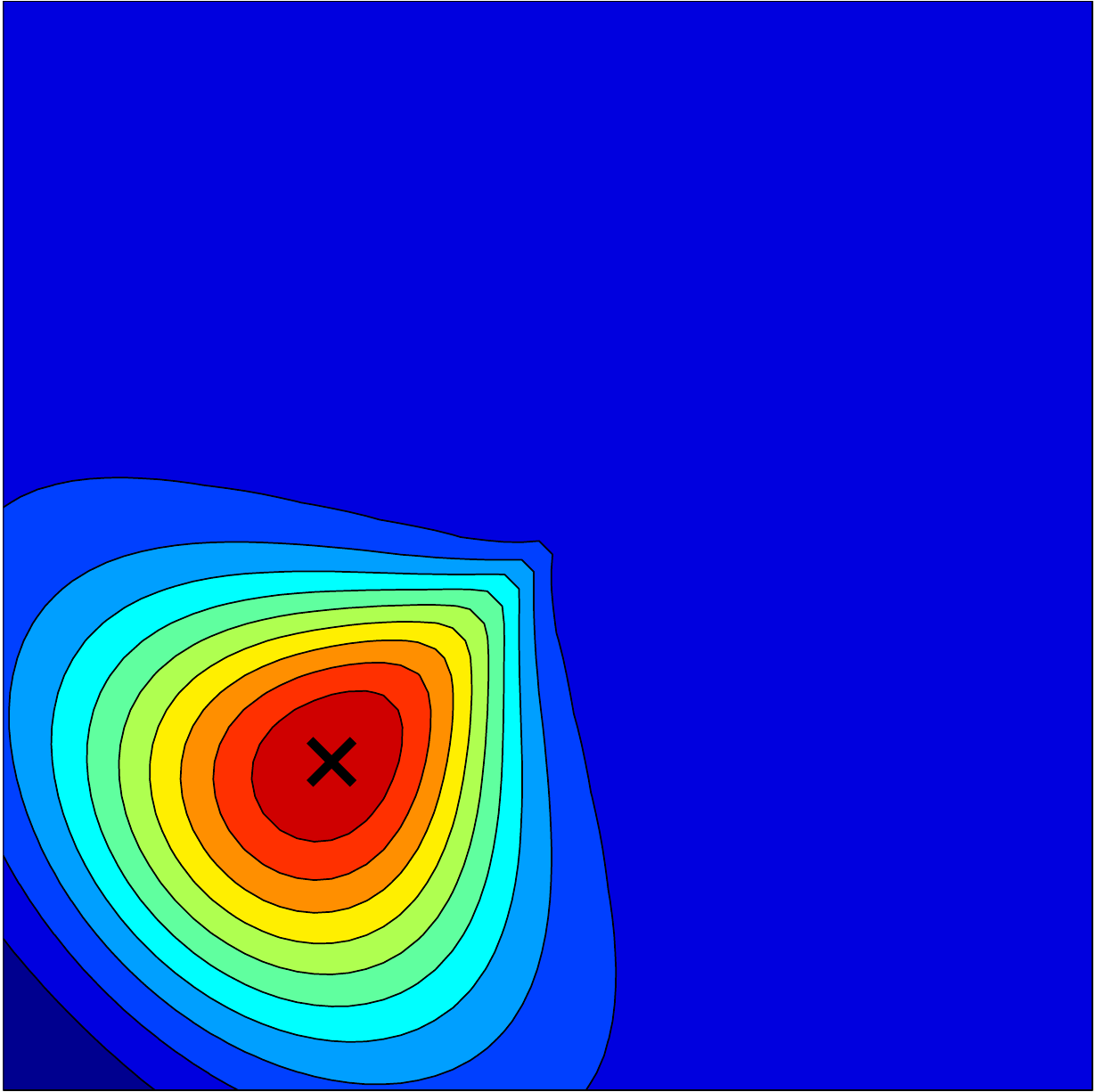}&
   \includegraphics[height=75pt]{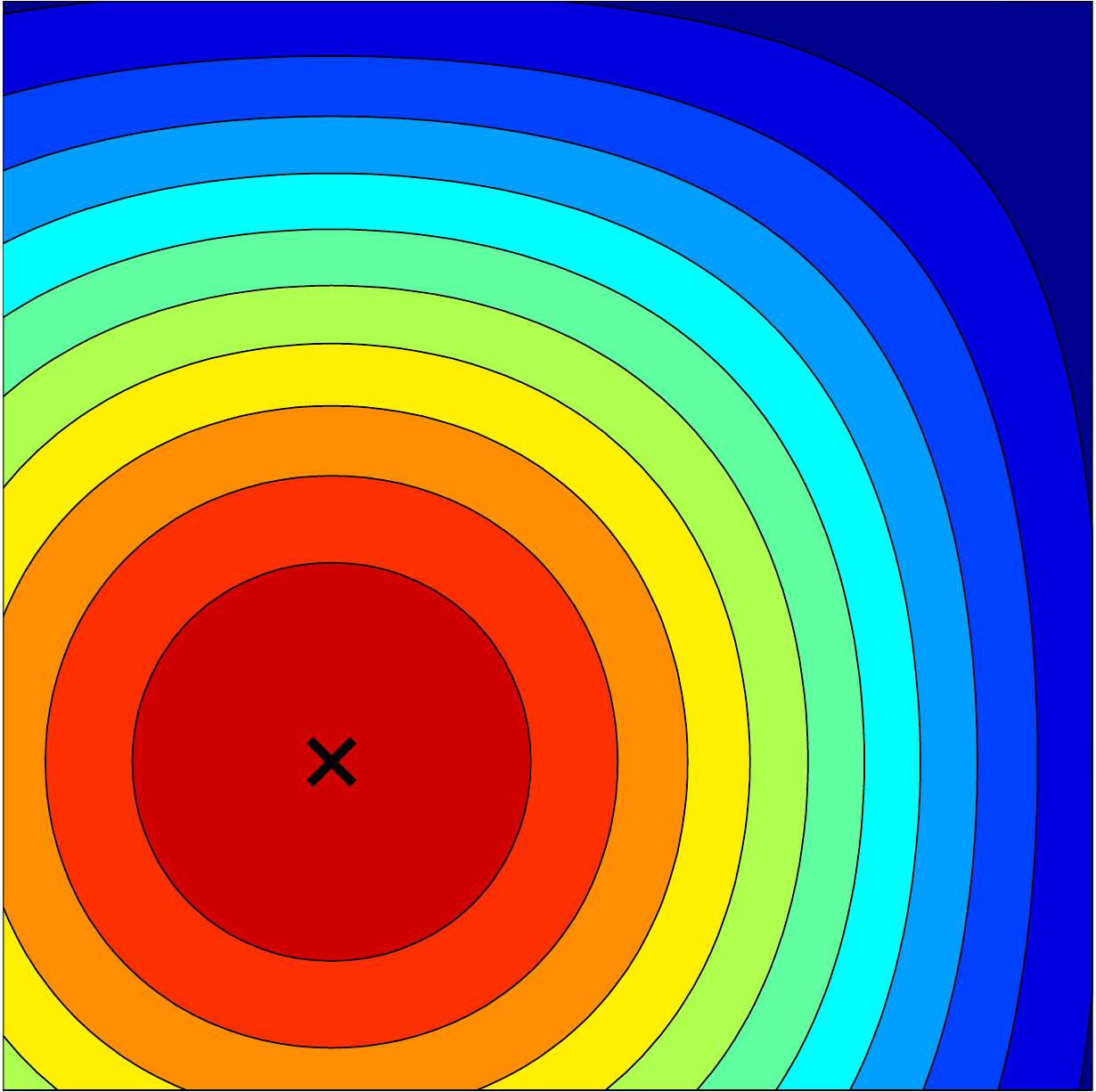}&
   \includegraphics[height=75pt]{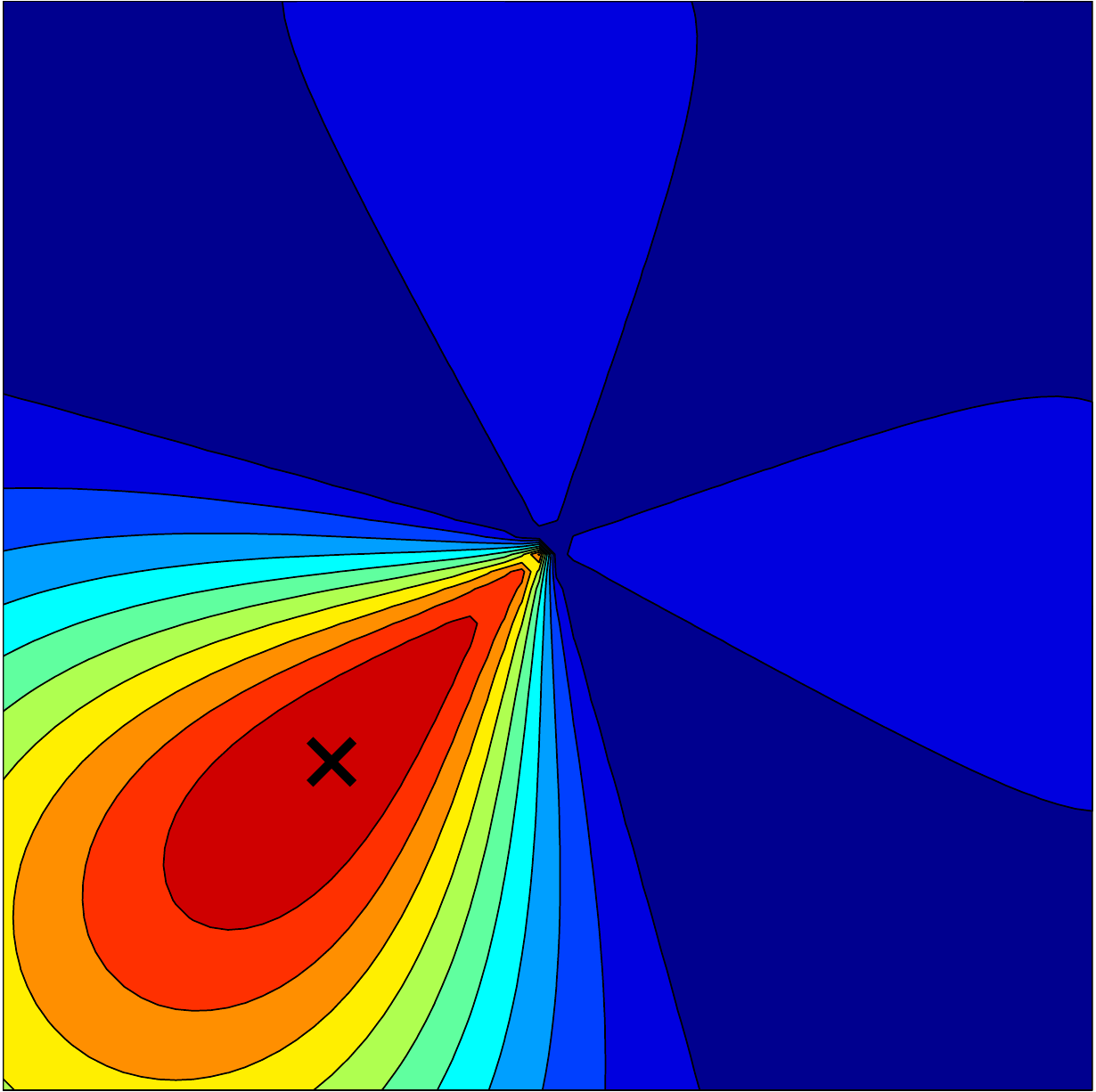}
   \vspace{15pt}\\
   \rotatebox{90}{$\hspace{15pt}\Delta\theta = \nicefrac{\pi}{8}$} &
   \includegraphics[height=75pt]{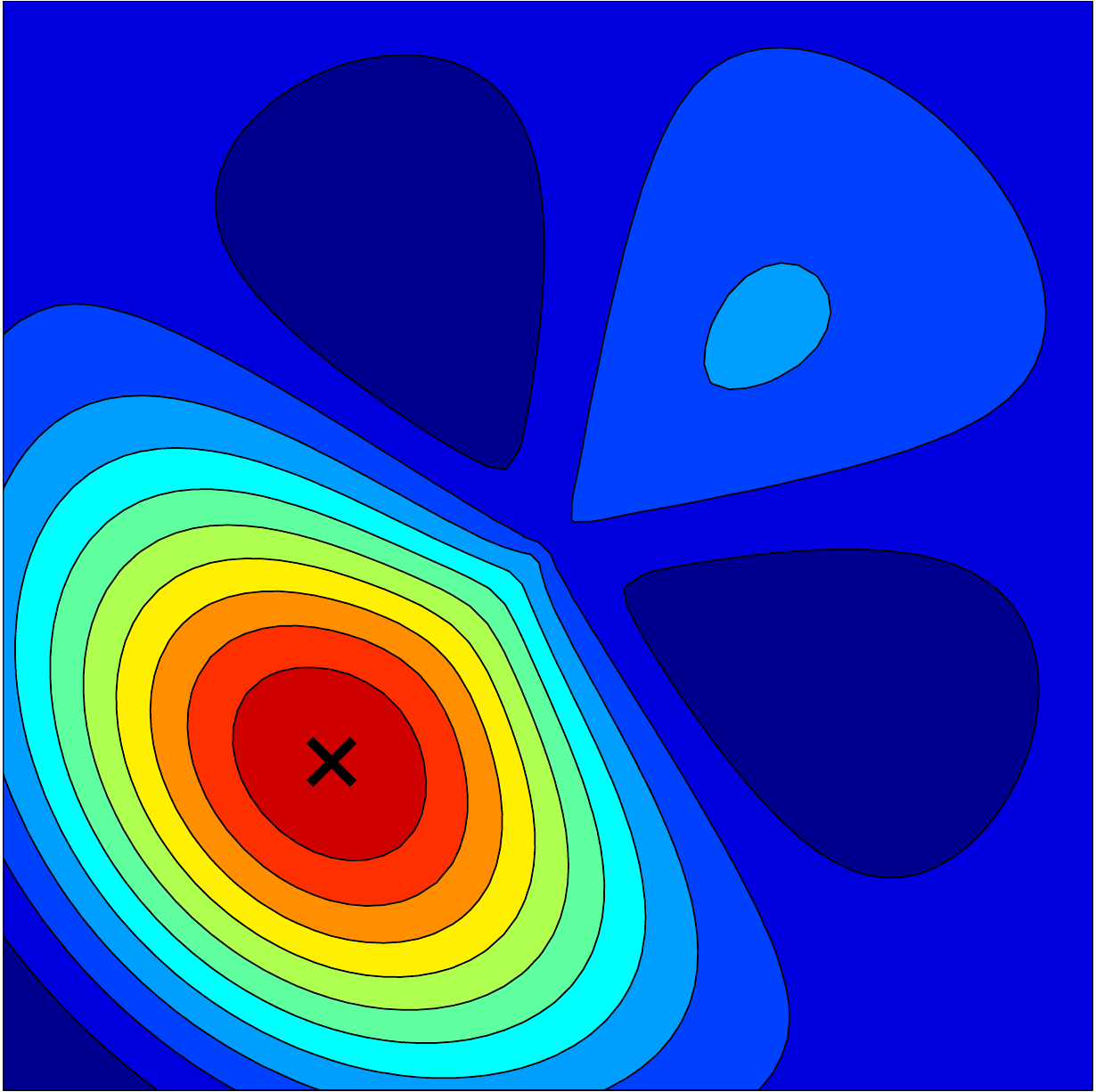}&
   \includegraphics[height=75pt]{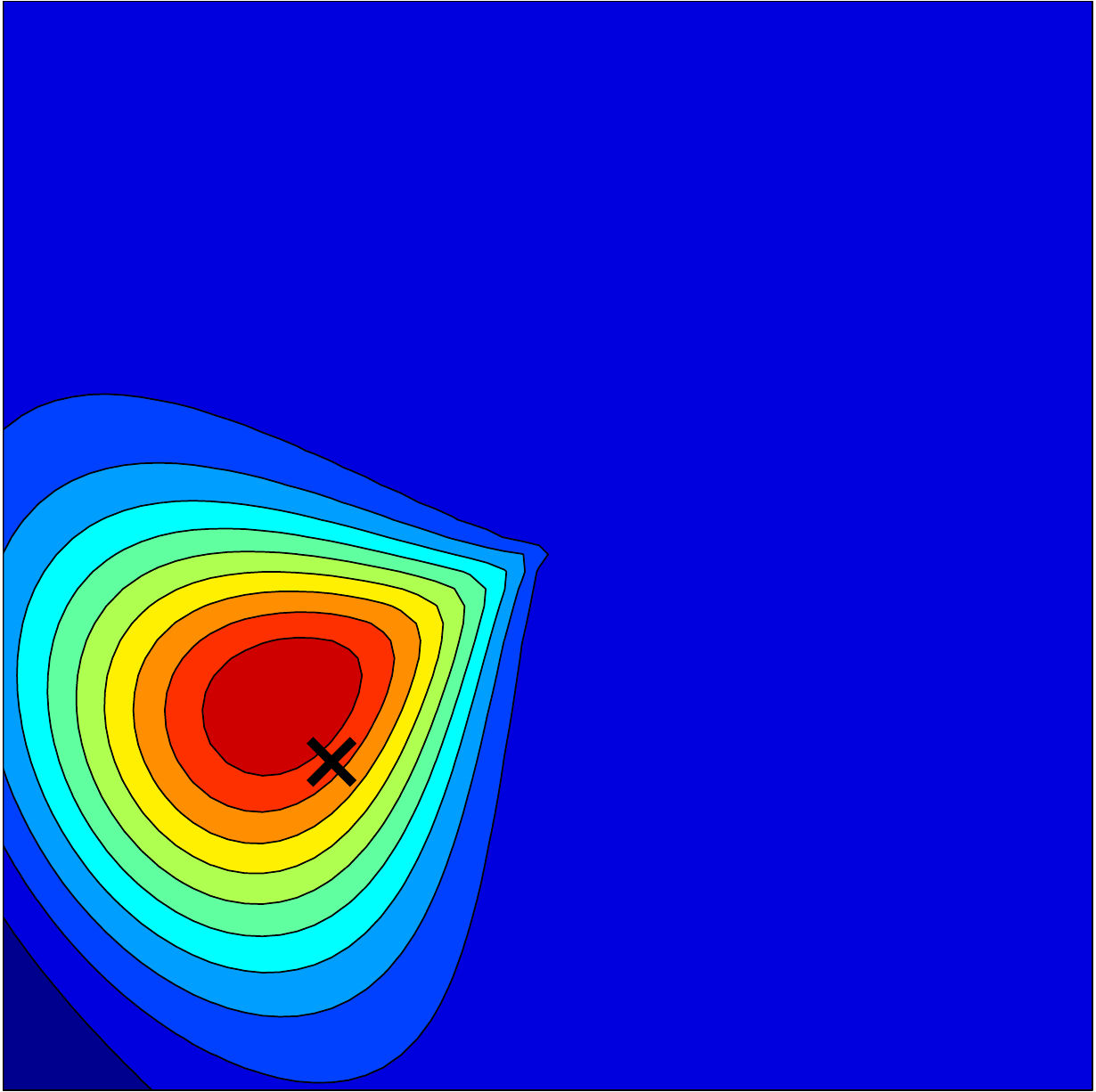}&
   \includegraphics[height=75pt]{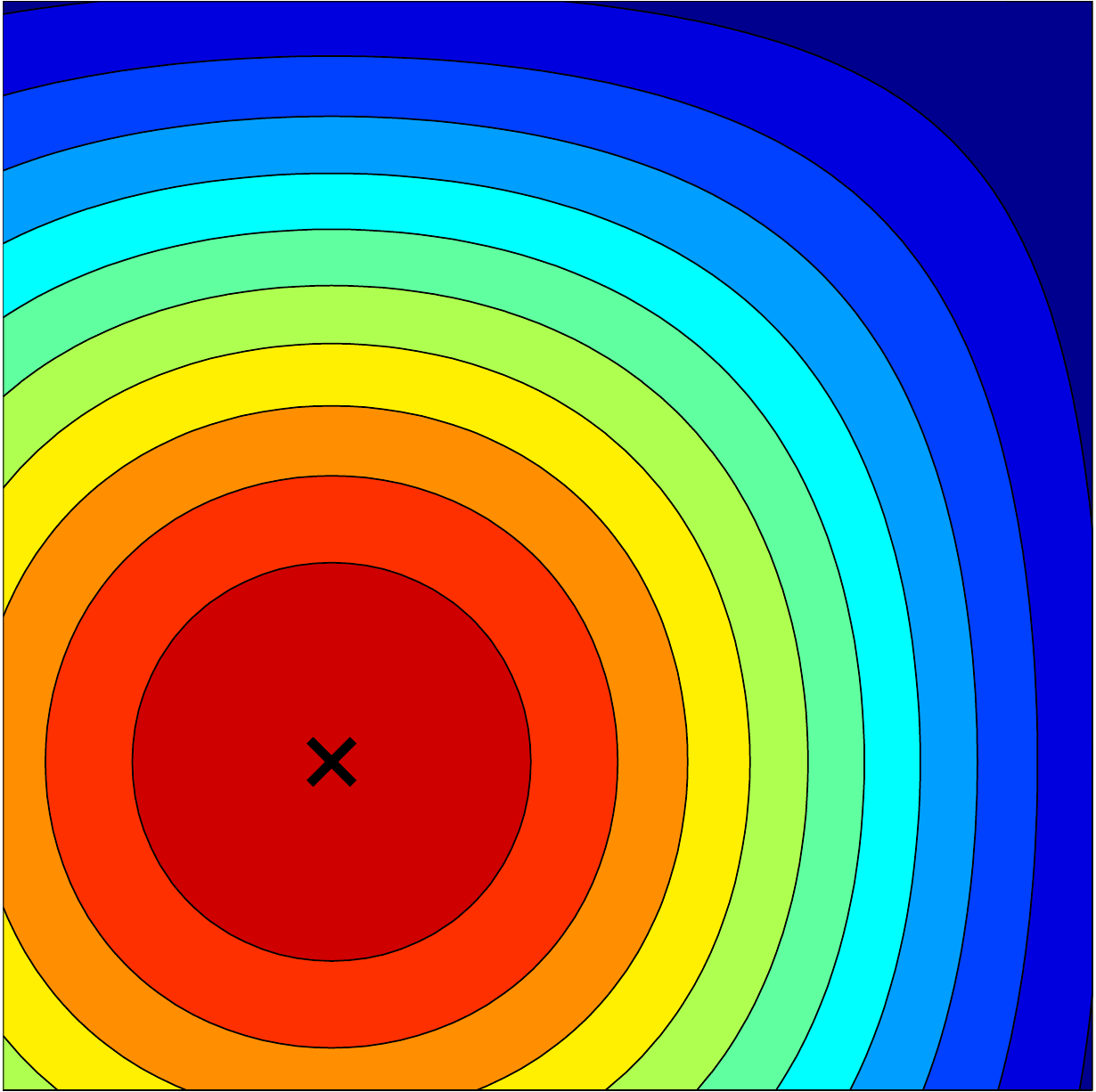}&
   \includegraphics[height=75pt]{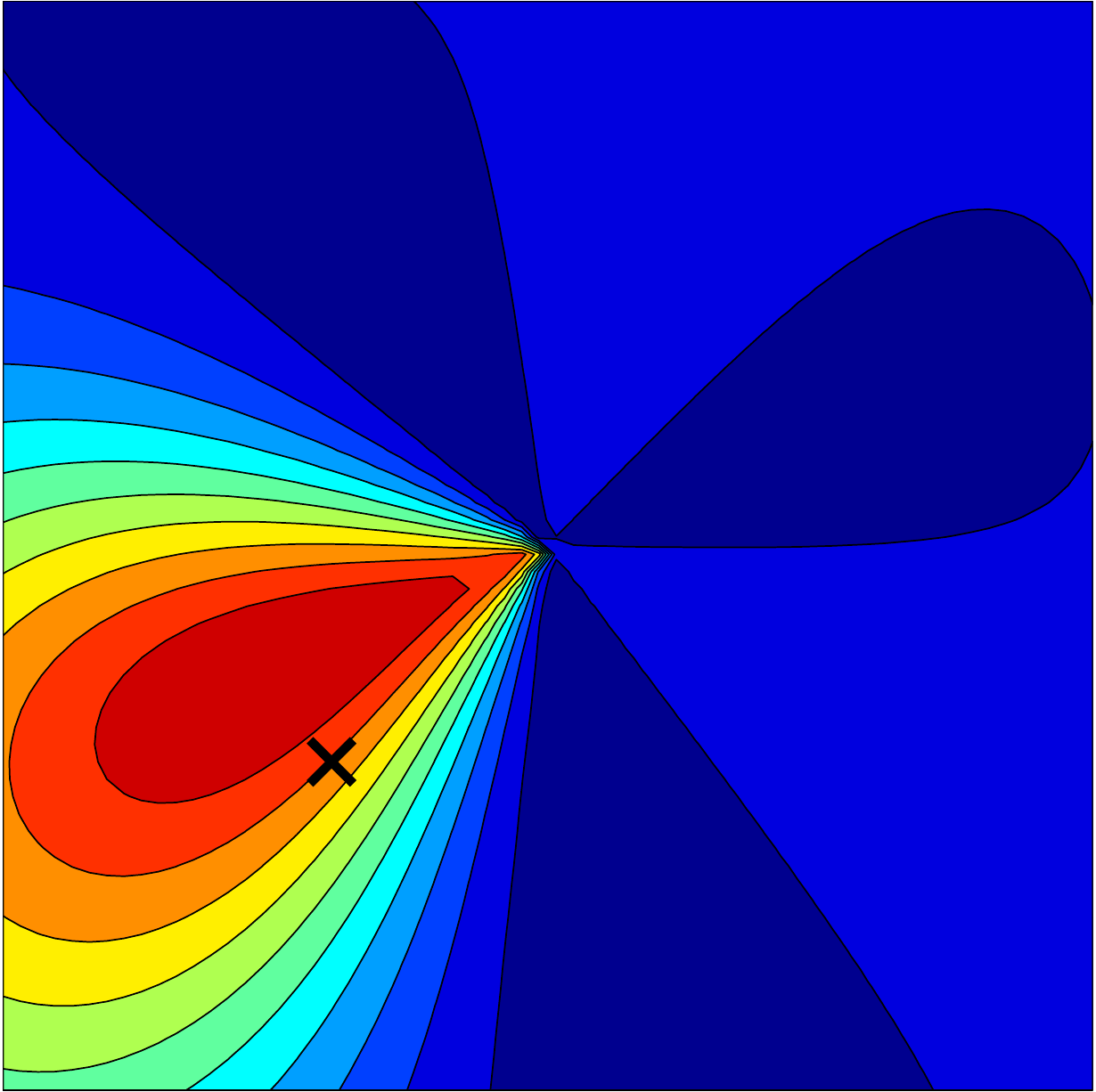}
   \vspace{10pt}\\
   & \kp\hspace{-2pt}\kr\hspace{-2pt}\kt &  \kp\hspace{-2pt}\kr\hspace{-2pt}\ktt & \kx\hspace{-2pt}\ky\hspace{-2pt}\kt & \kx\hspace{-2pt}\ky\hspace{-2pt}\ktt
\end{tabular}
\vspace{10pt}
\caption{Patch maps for different parametrizations and kernels. We present two parametrizations in polar and two in cartesian coordinates, with absolute or relative gradient angle for each one. $\Delta\theta$ is fixed and pixel \p is shown with ``$\times$''. At the bottom of each column the kernels (patch similarity) approximated are shown.\protect\footref{footfig}\label{fig:parametrizations}}
\end{figure*}

\vspace{5pt}
For example, in Figure~\ref{fig:parametrizations} patch maps for different kernels are shown. The position of \p is denoted by $\times$ symbol. The value of $\p_\theta = 0$ and $\q_\theta = 0$ for all spatial locations of \q in the top row and $\q_\theta = -\pi/8$ in the bottom row.
The visualization shows the discontinuity of the pixel similarity impact of the $\V_{\prma}$ descriptor near the center of the patch. This is caused by the polar coordinate system where a small difference in the position near the origin causes large difference in $\phi$ and $\tilde{\theta}$. Also in the bottom row we see that using the relative gradient direction $\tilde{\theta}$ allows to compensate for imprecision caused by small patch rotation, \ie the most similar pixel is not the one at the location of \p with different $\tilde{\theta}$, but a rotated pixel with more similar value of  $\tilde{\theta}$. Finally, we observe that the kernel parametrized by Cartesian coordinates and absolute angle of the gradient ($\V_{\prmb}$, third column) is insensitive to small translations, \ie feature point displacement.

\vspace{5pt}
We additionally construct patch maps in the case of descriptor post-processing by a linear transformation, \eg descriptor whitening.
Now the contribution of a pixel pair is given by
\vspace{5pt}
\begin{align}
\hat{\M}(\p,\q) &= (A^\top(\map(\p)-\mu))^\top (A^\top(\map(\q)-\mu))\\
                &= (\map(\p)-\mu)^\top AA^\top(\map(\q)-\mu)\\
					 &= \map(\p)^\top AA^\top \map(\q) - \map(\p)^\top AA^\top\mu - \map(\q)^\top AA^\top\mu + \mu^\top AA^\top \mu.
\end{align}
%
The last term is constant and can be ignored, while if $A$ is a rotation matrix then only shifting by $\mu$ affects the similarity. After the transformation, the similarity is no longer shift-invariant. The non-linear post-processing, such as power-law normalization or simple \l2 normalization cannot be visualized, as it acts after the pixel aggregation\footnote{Details are omitted due to lack of space.}.

\begin{figure*}
\centering
\begin{tabular}{@{\tsp}c@{\hspace{-2pt}}ccc}
   \raisebox{40pt}{\begin{tabular}{l}$\Delta\theta = 0$\\ \pt=0\\ \qt=0\end{tabular}} &
   \includegraphics[height=95pt]{fig/patchsim/aa2r_rhofull_featuremap/p_20_20_d_0_0_.pdf}&
   \includegraphics[height=95pt]{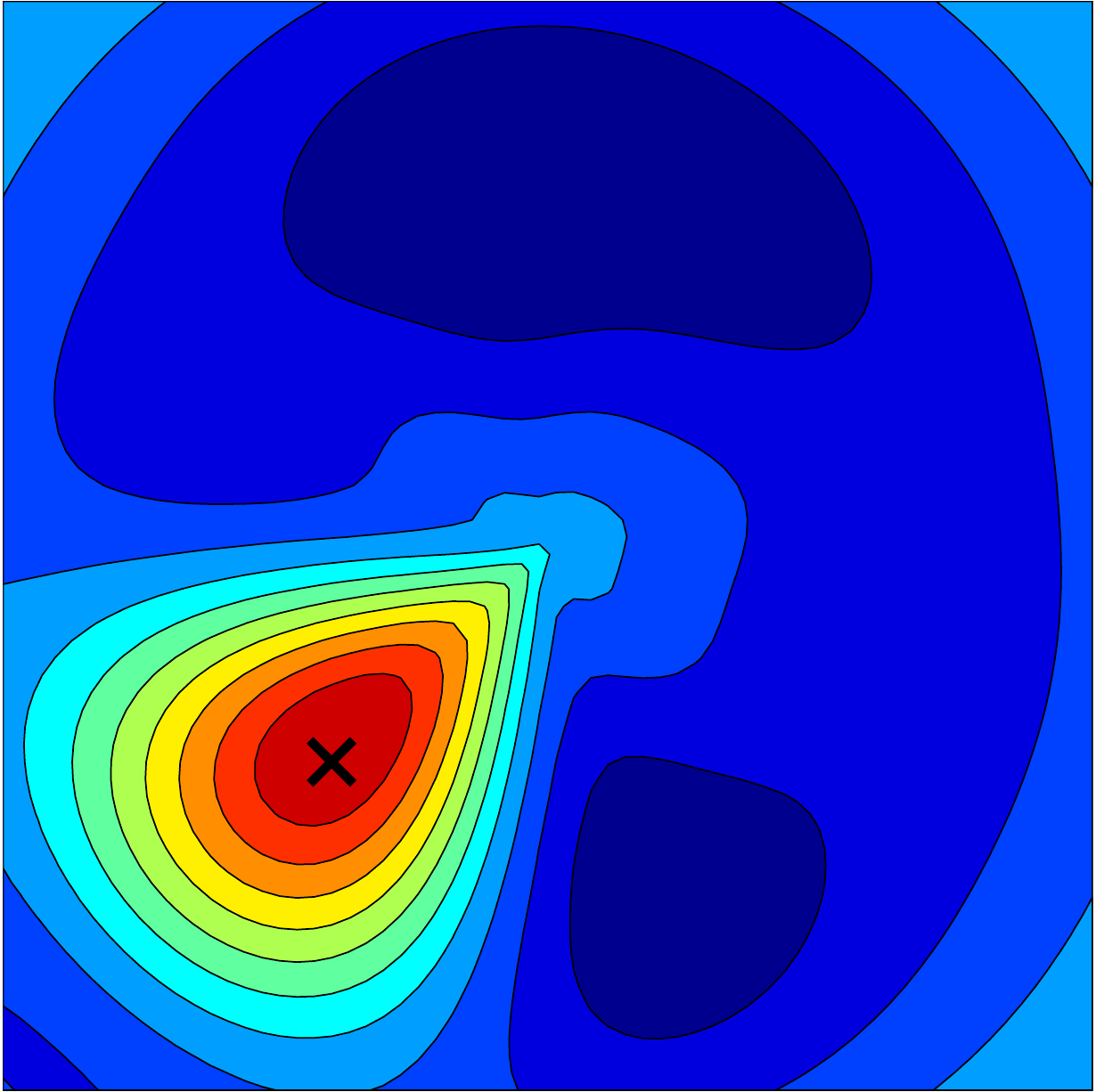}&
   \includegraphics[height=95pt]{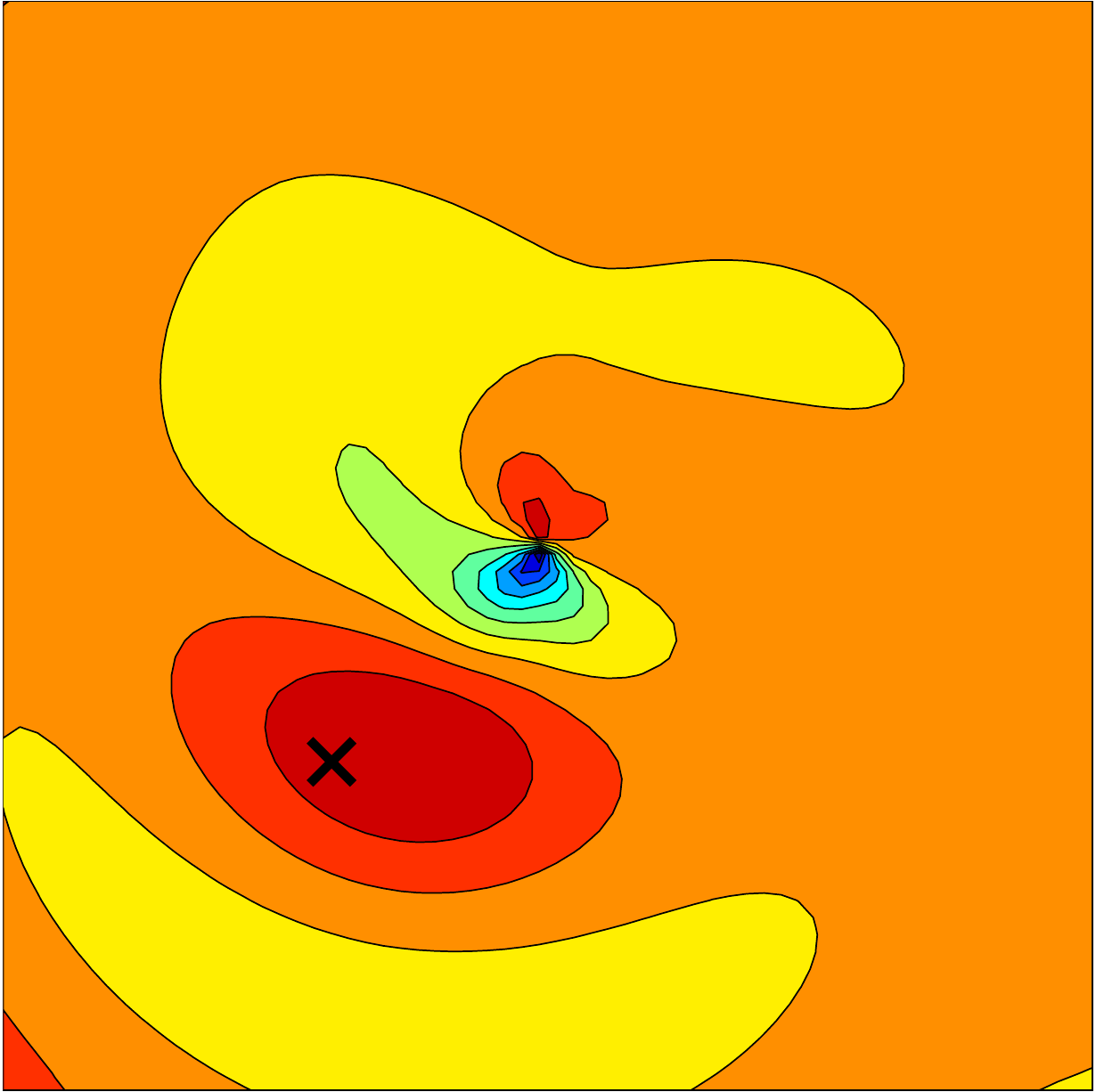} \vspace{15pt}\\
   \raisebox{40pt}{\begin{tabular}{l}$\Delta\theta = \nicefrac{\pi}{8}$\\ \pt=0\\ $\qt=-\nicefrac{\pi}{8}$\end{tabular}} &
   \includegraphics[height=95pt]{fig/patchsim/aa2r_rhofull_featuremap/p_20_20_d_0_39.pdf}&
   \includegraphics[height=95pt]{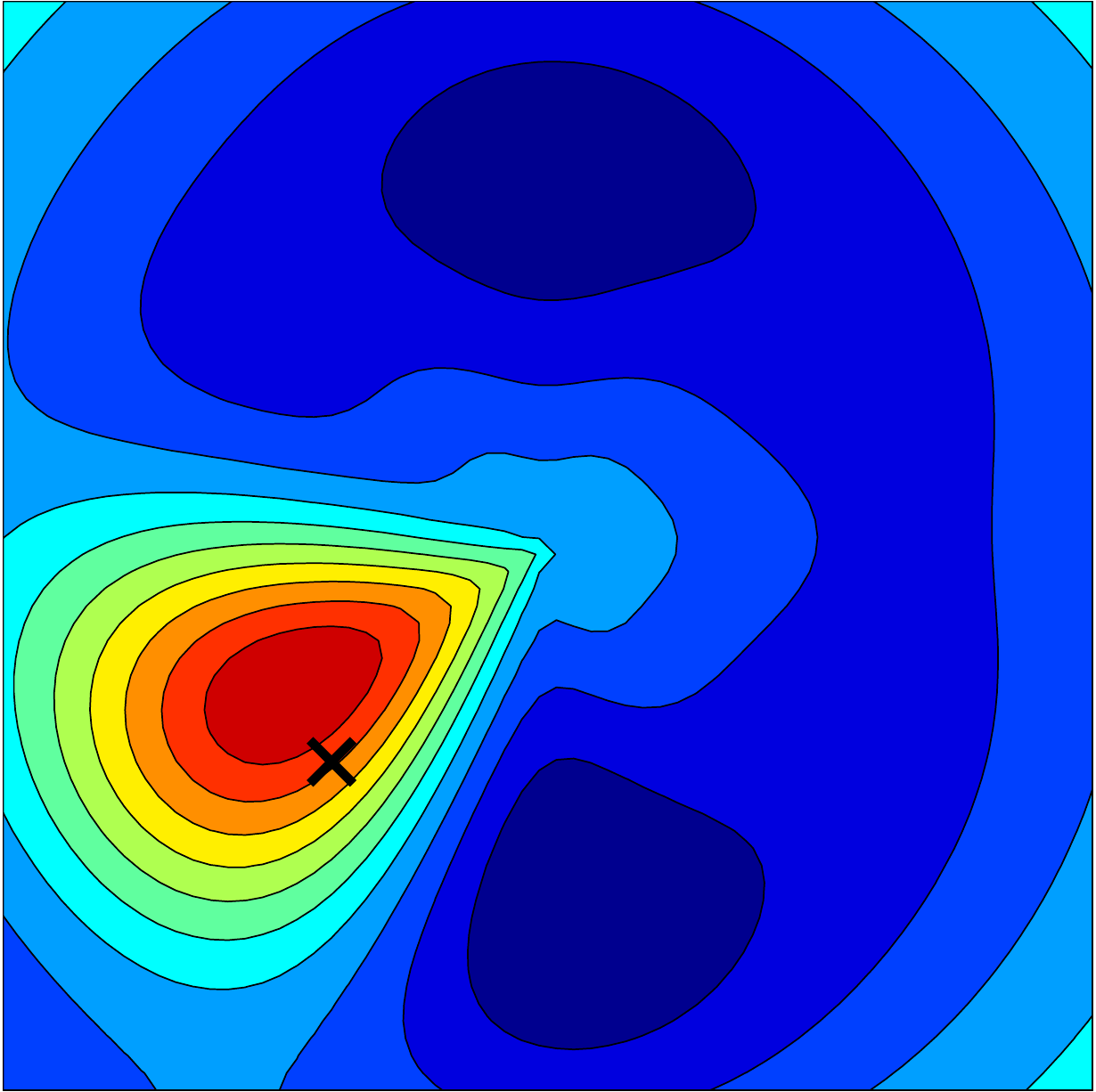}&
   \includegraphics[height=95pt]{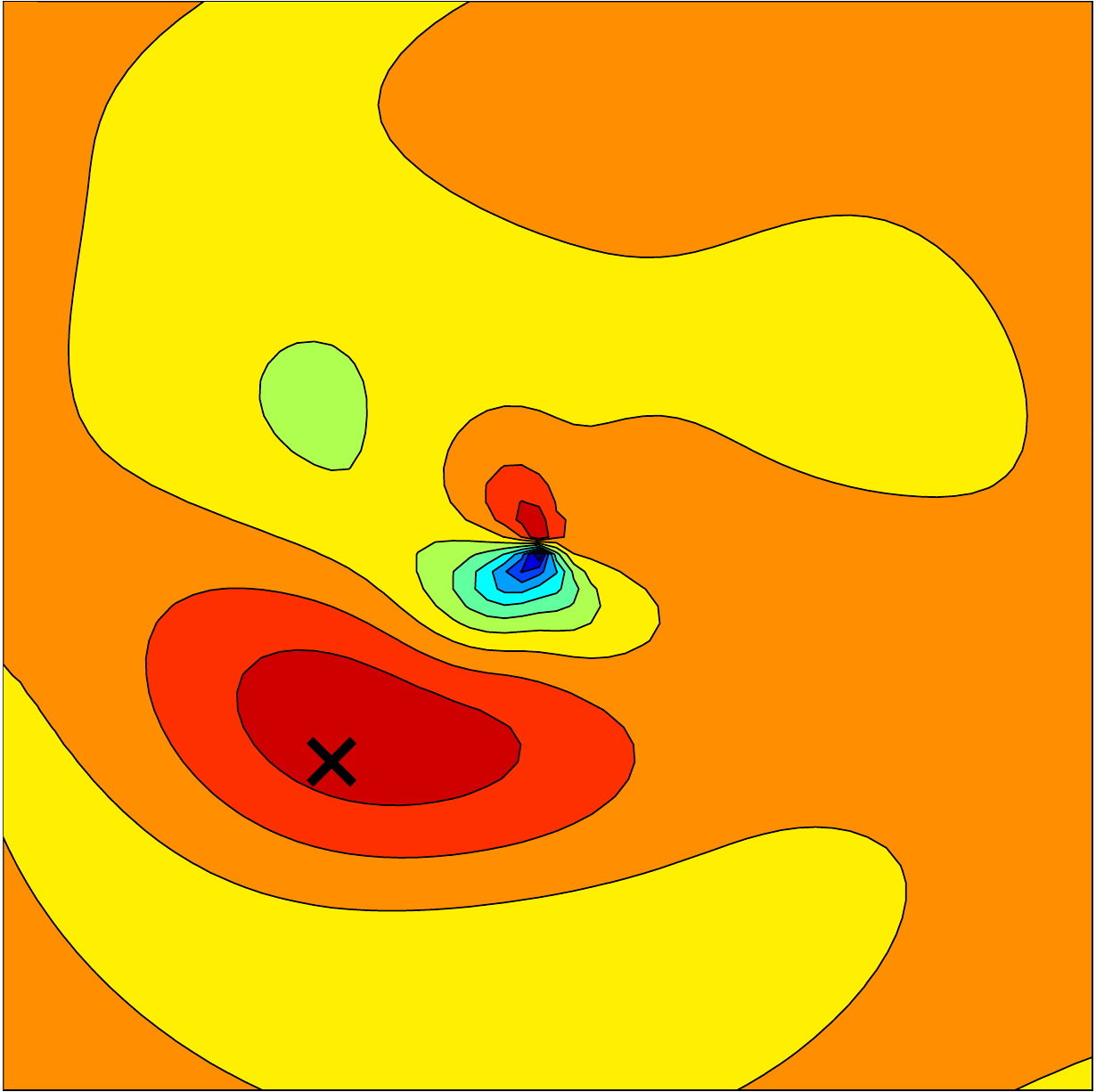} \vspace{10pt}\\
   & No transformation & PCA & Learned whitening (LW)
\end{tabular}
\vspace{10pt}
\caption{Patch maps for \prma parametrization and kernels. $\Delta\theta$ is fixed by choosing fixed values for \pt and \qt. Pixel \p is shown with ``$\times$''. Three different cases are shown: without transformation, with PCA transformation and with transformation by supervised whitening.\protect\footref{footfig}
\label{fig:baseline}}
\end{figure*}

\vspace{5pt}
Figure~\ref{fig:baseline} we shows patch maps for $\V_{\prma}$ in the case of PCA or LW post-processing. PCA is shown to have some small effect on the similarity, while LW significantly changes the derived shape. It implicitly affects the shape of the kernels used; observe that the kernels go wider in the circular direction.

\vspace{10pt}
\textbf{Combining kernel descriptors.}
We propose to take advantage of both parametrizations $\V_{\prma}$ and $\V_{\prmb}$, by summing their contribution. This is performed by simple concatenation of the two descriptors. Finally, whitening is jointly learned and dimensionality reduction is performed.

\begin{figure*}[t!]
\vspace{10pt}
\footnotesize
\begin{tabular}{@{\ssp}c@{\ssp}c@{\ssp}c@{\ssp}c@{\ssp}c@{\hspace{-1pt}}c@{\ssp}}
   \includegraphics[height=58pt]{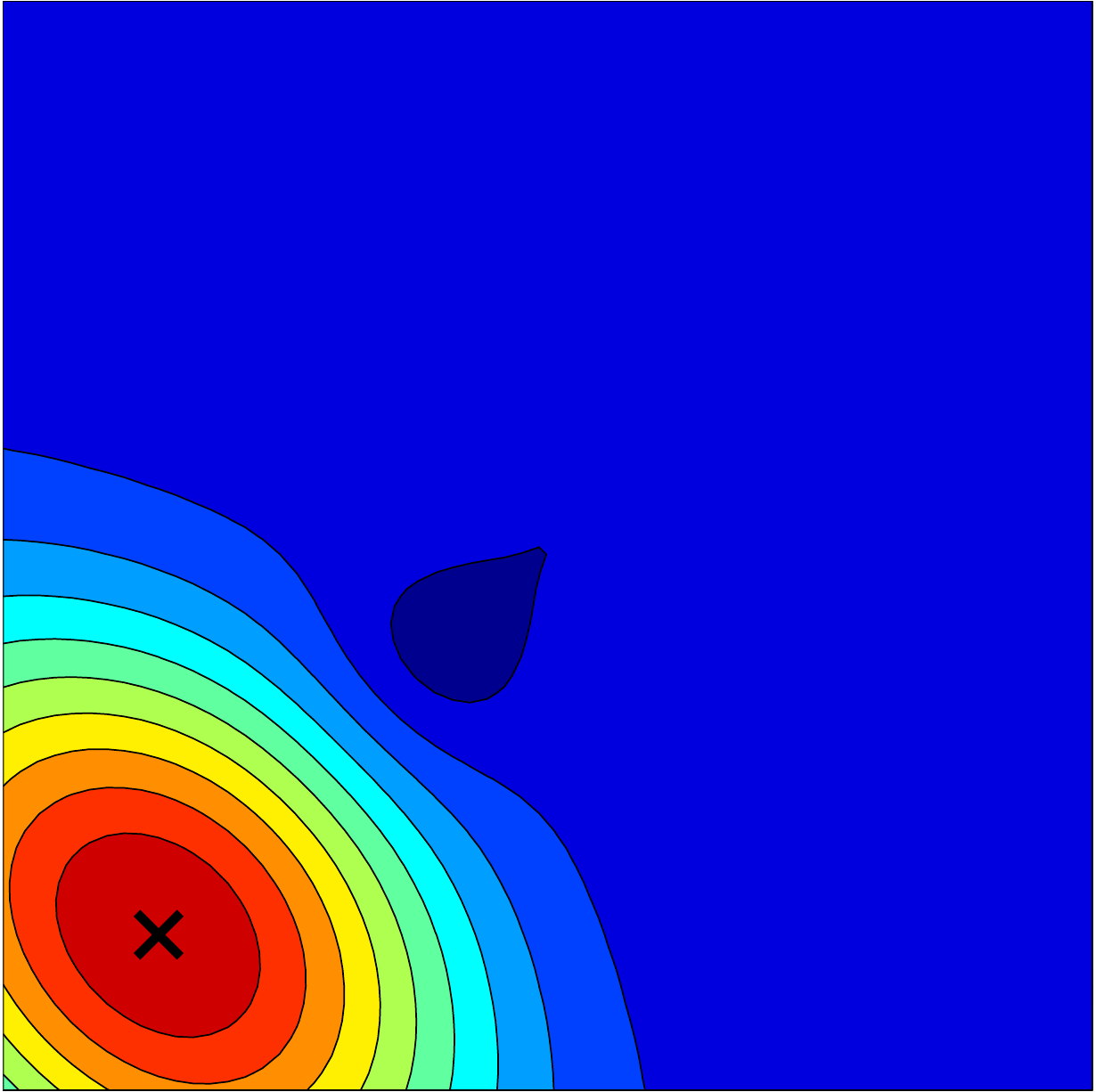}&
   \includegraphics[height=58pt]{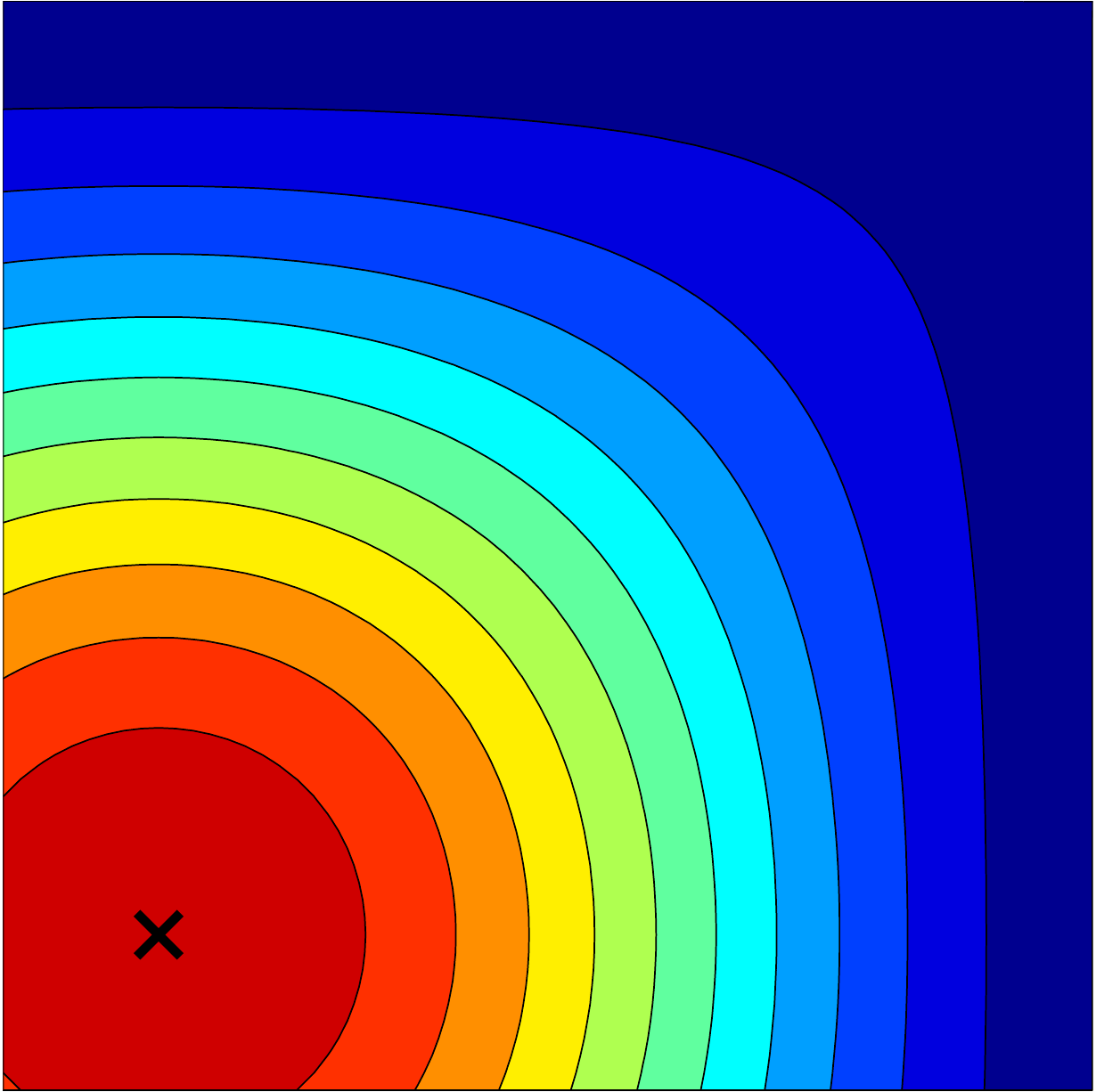}&
   \includegraphics[height=58pt]{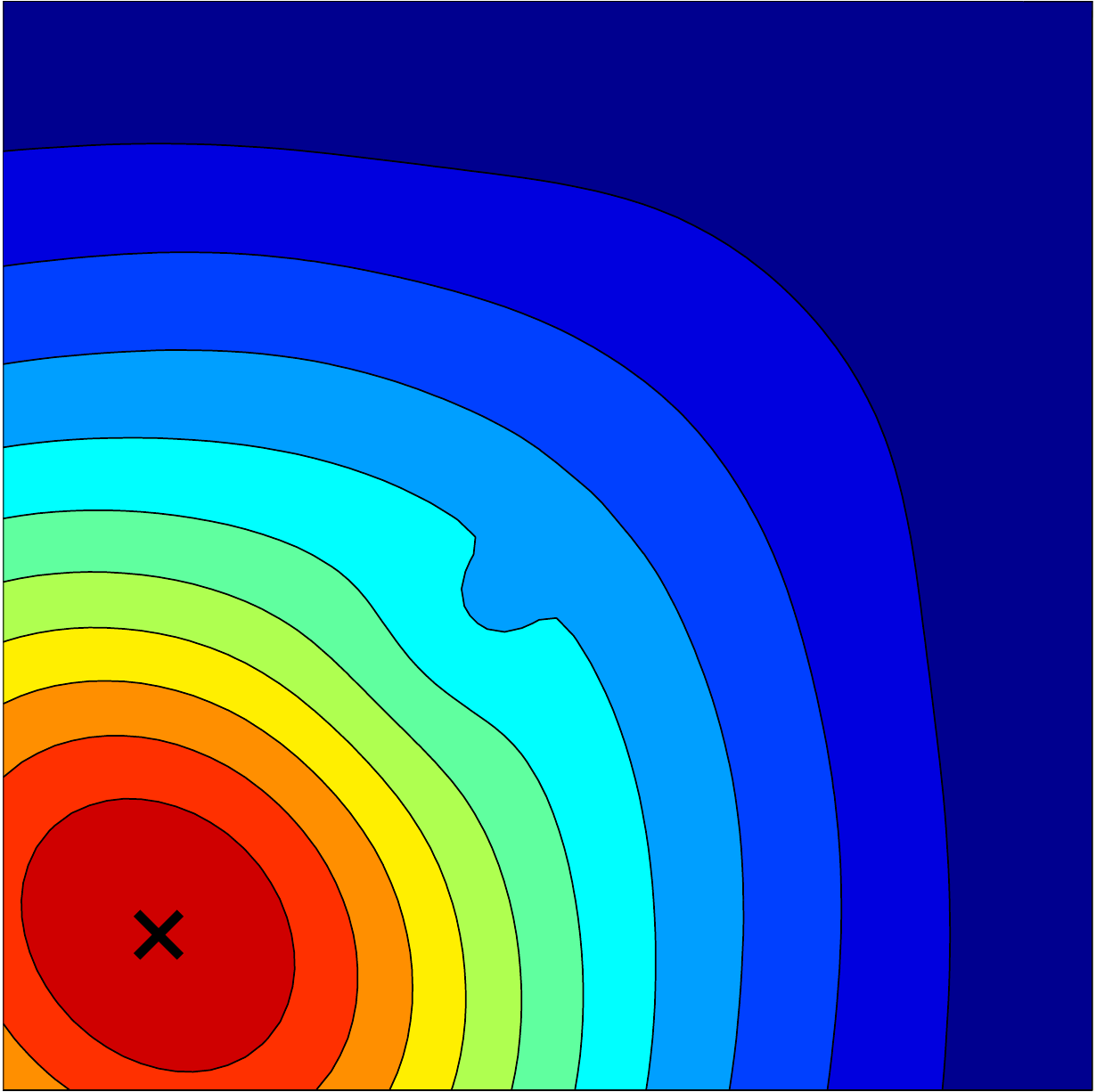}&
   \includegraphics[height=58pt]{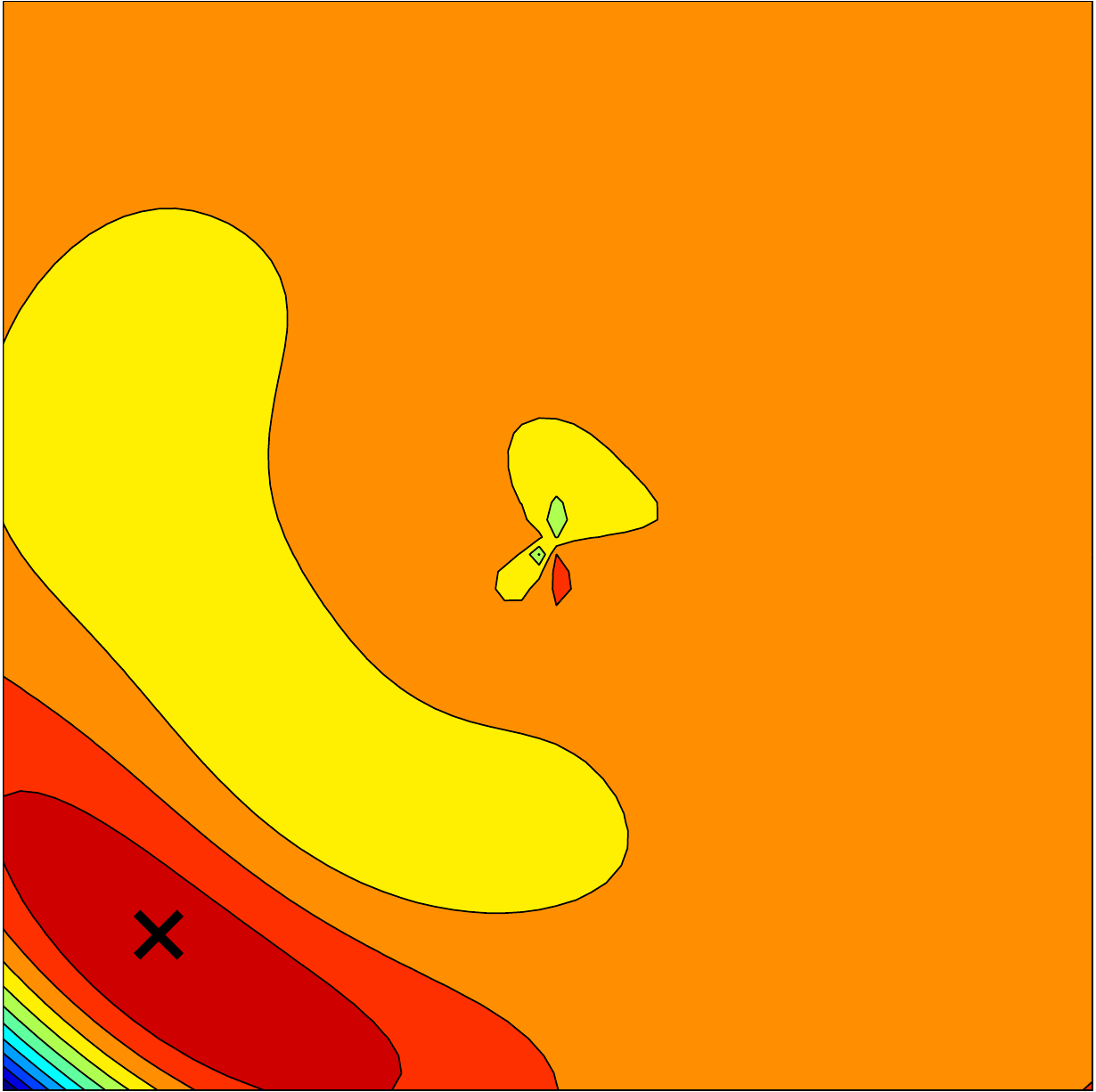}&
   \includegraphics[height=58pt]{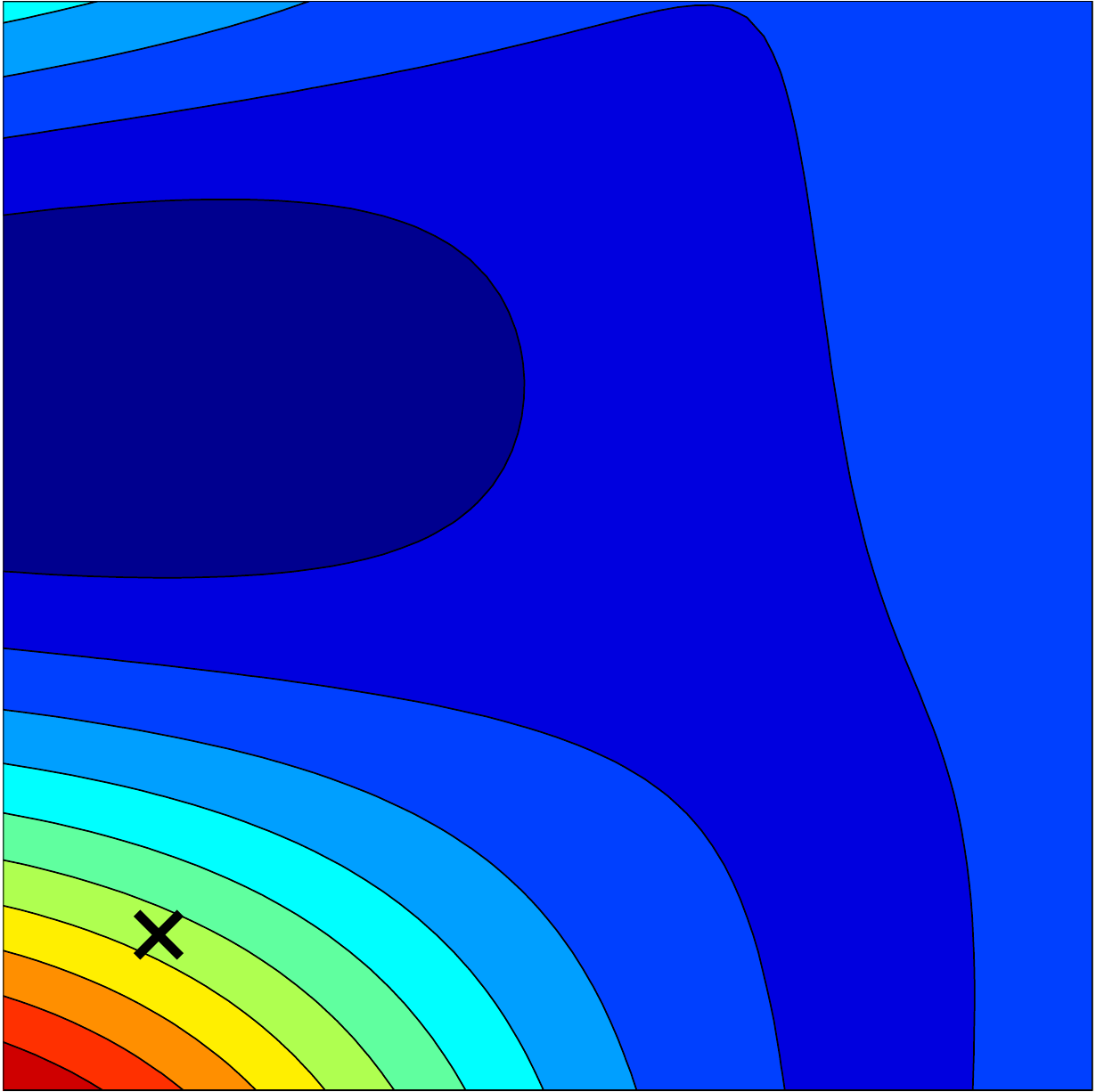}&
   \includegraphics[height=58pt]{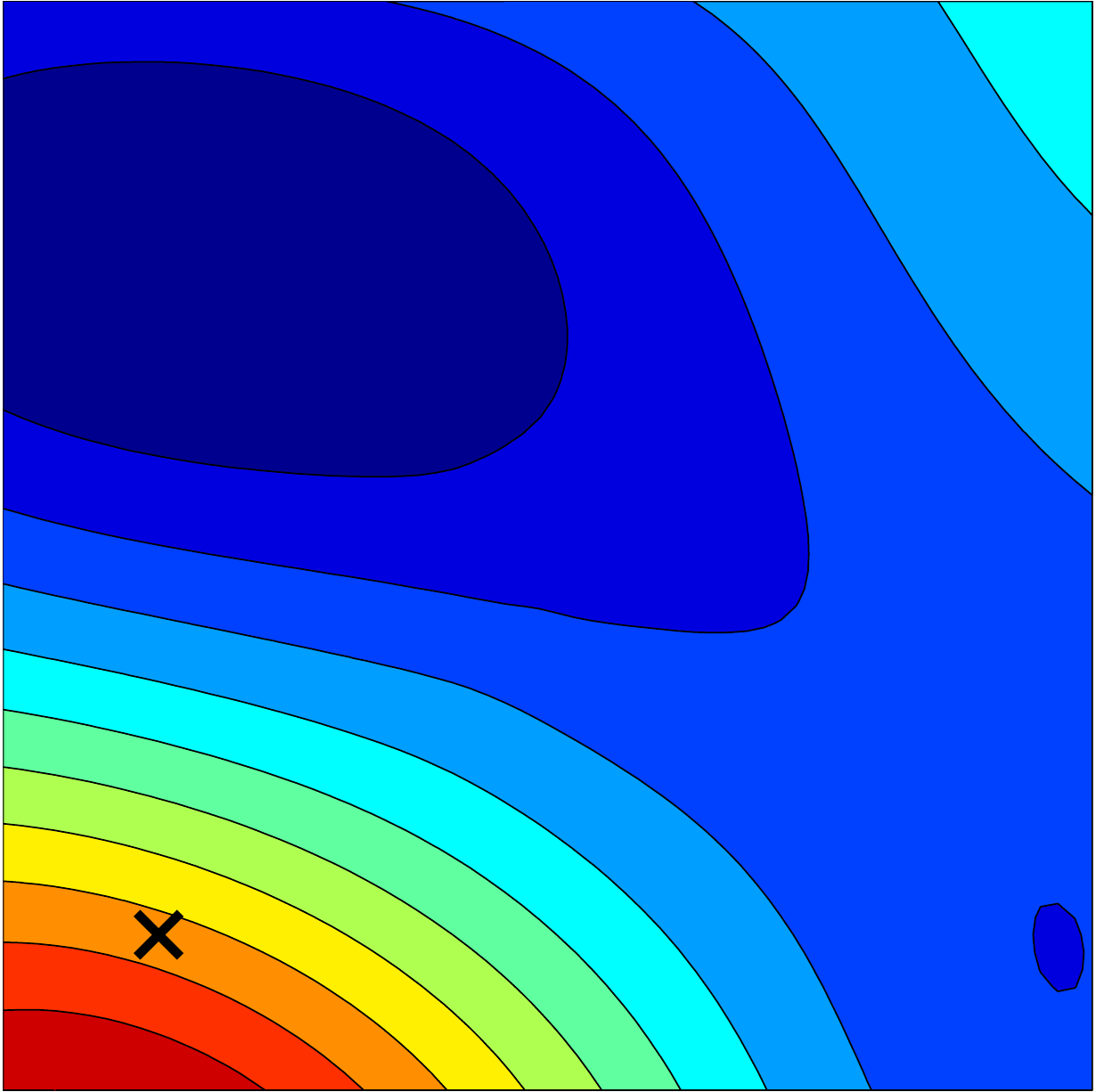} \vspace{8pt}\\
   \includegraphics[height=58pt]{fig/patchsim/aa2r_rhofull_featuremap/p_20_20_d_0_0_.pdf}&
   \includegraphics[height=58pt]{fig/patchsim/xya1_rhofull_smallkappa_featuremap/p_20_20_d_0_0_.pdf}&
   \includegraphics[height=58pt]{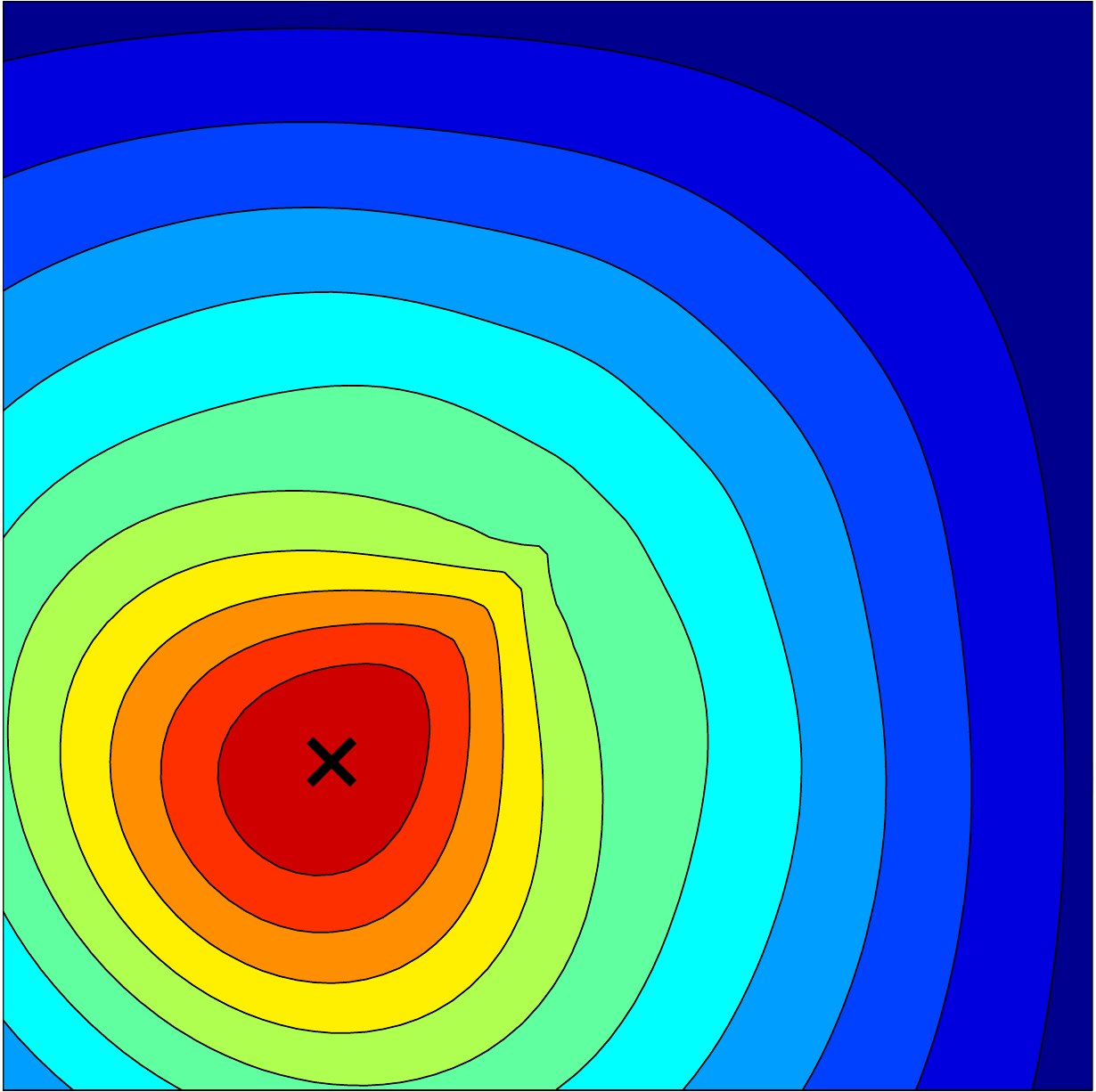}&
   \includegraphics[height=58pt]{fig/patchsim/aa2r_rhofull_featuremap_whitenl/p_20_20_d_0_0_.pdf}&
   \includegraphics[height=58pt]{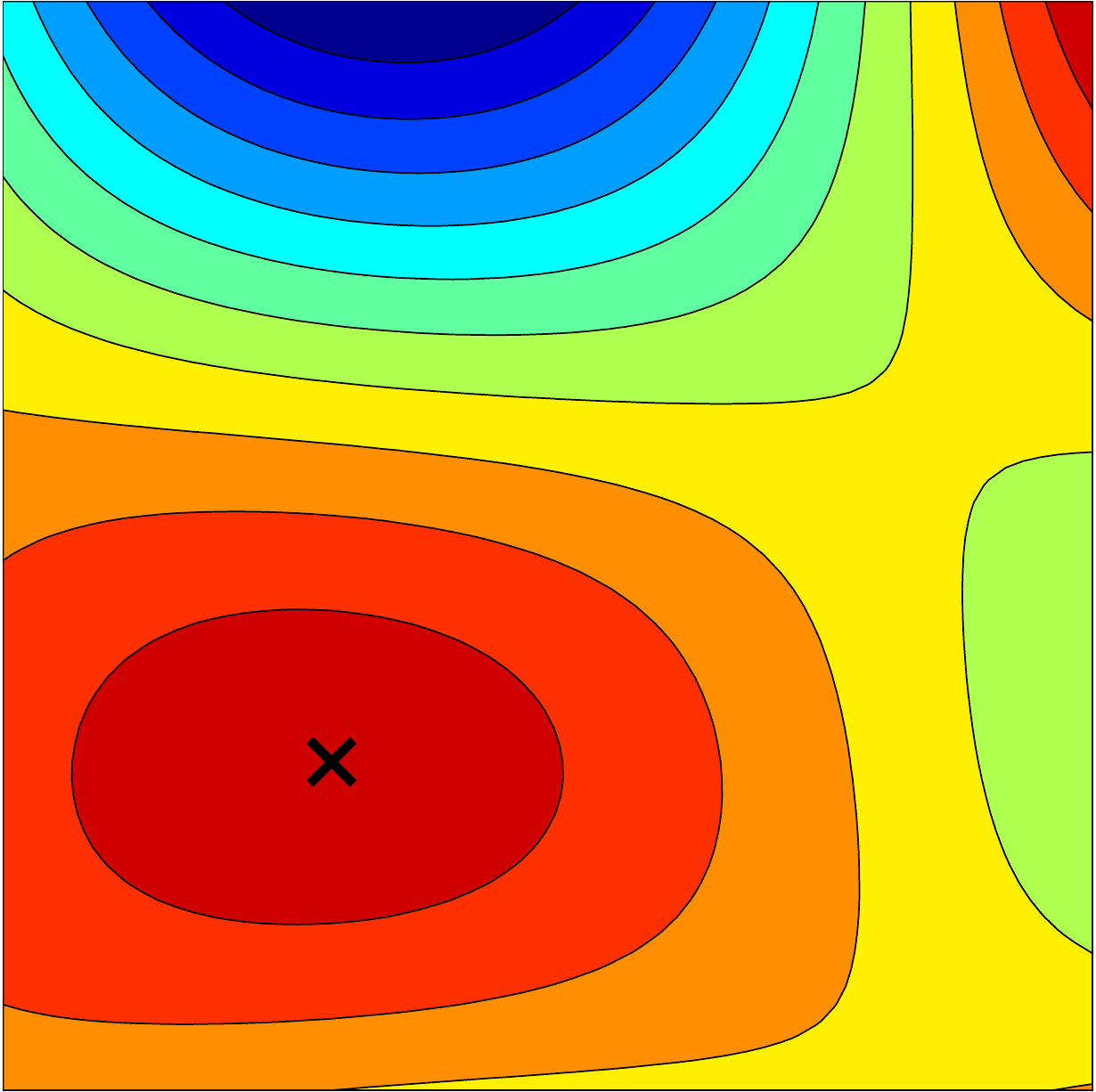}&
   \includegraphics[height=58pt]{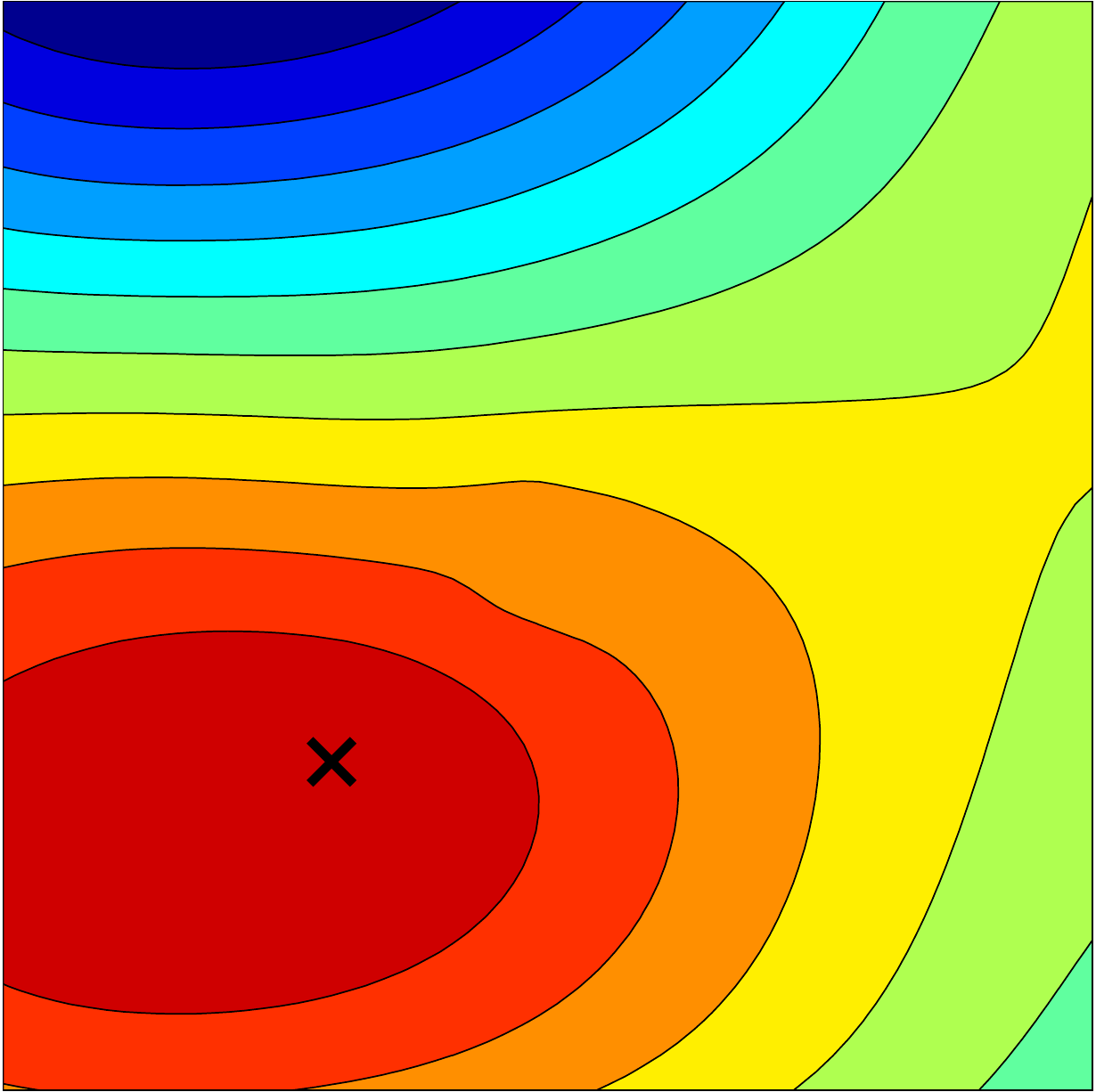} \vspace{8pt}\\
   \includegraphics[height=58pt]{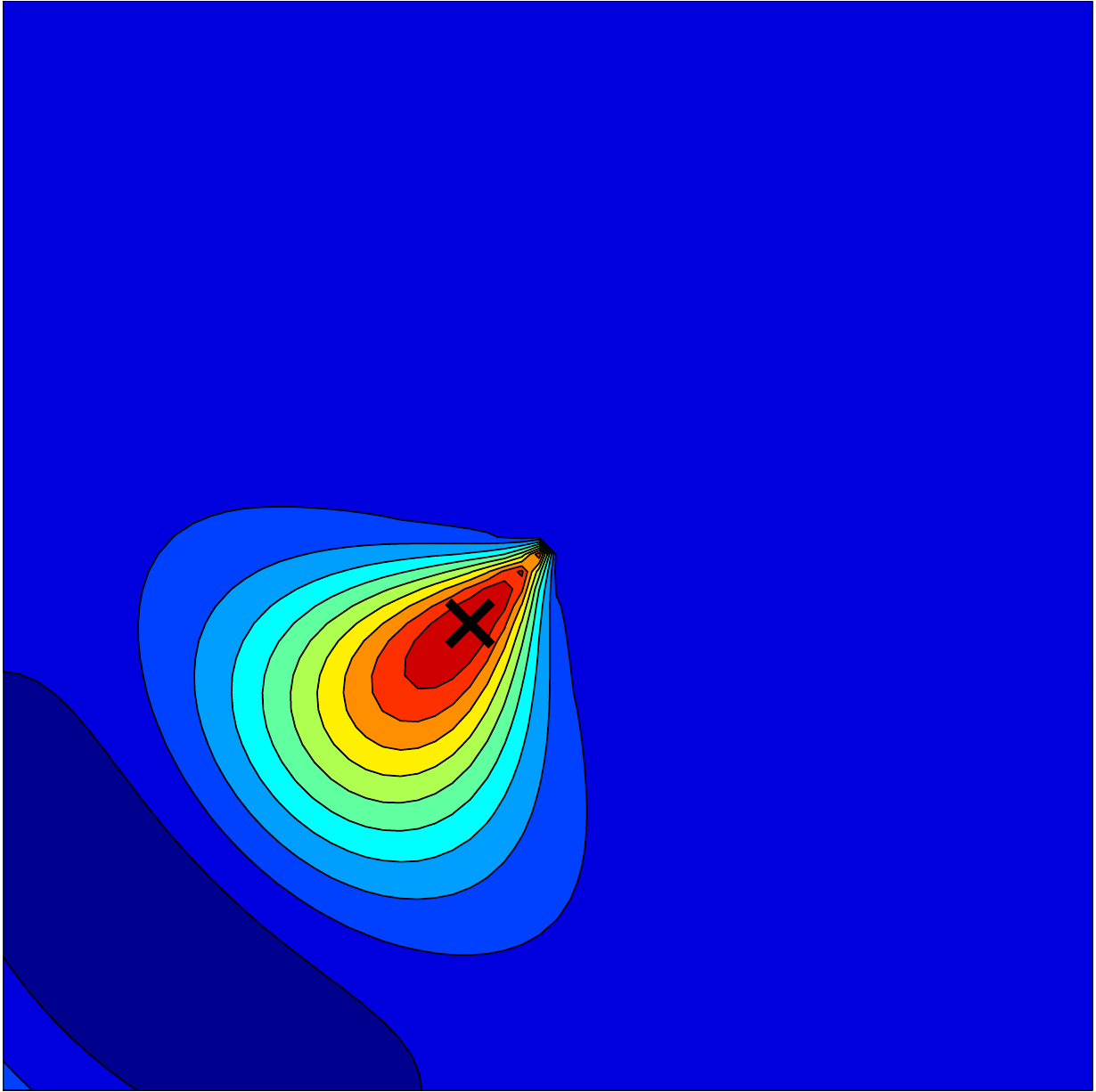}&
   \includegraphics[height=58pt]{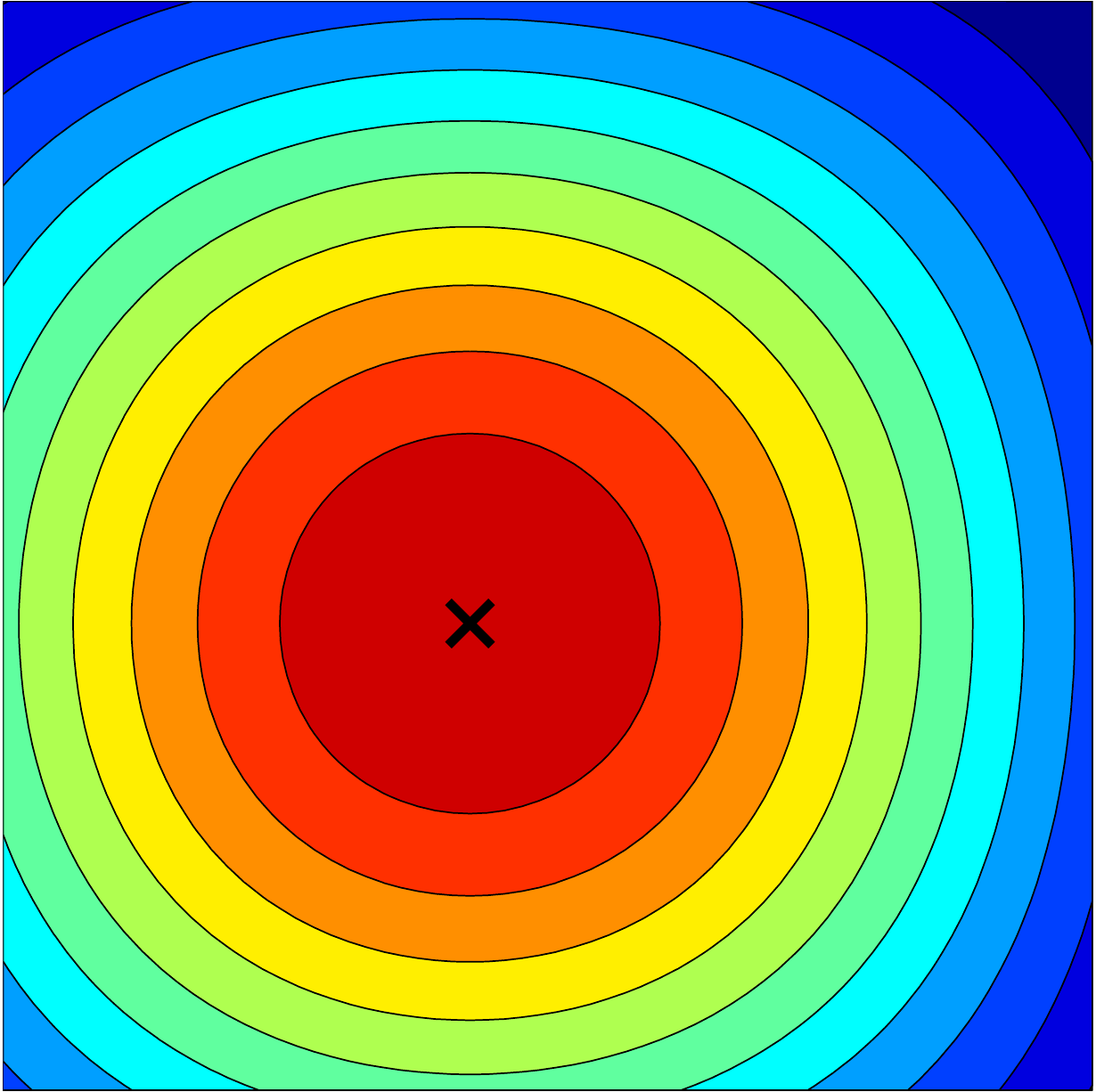}&
   \includegraphics[height=58pt]{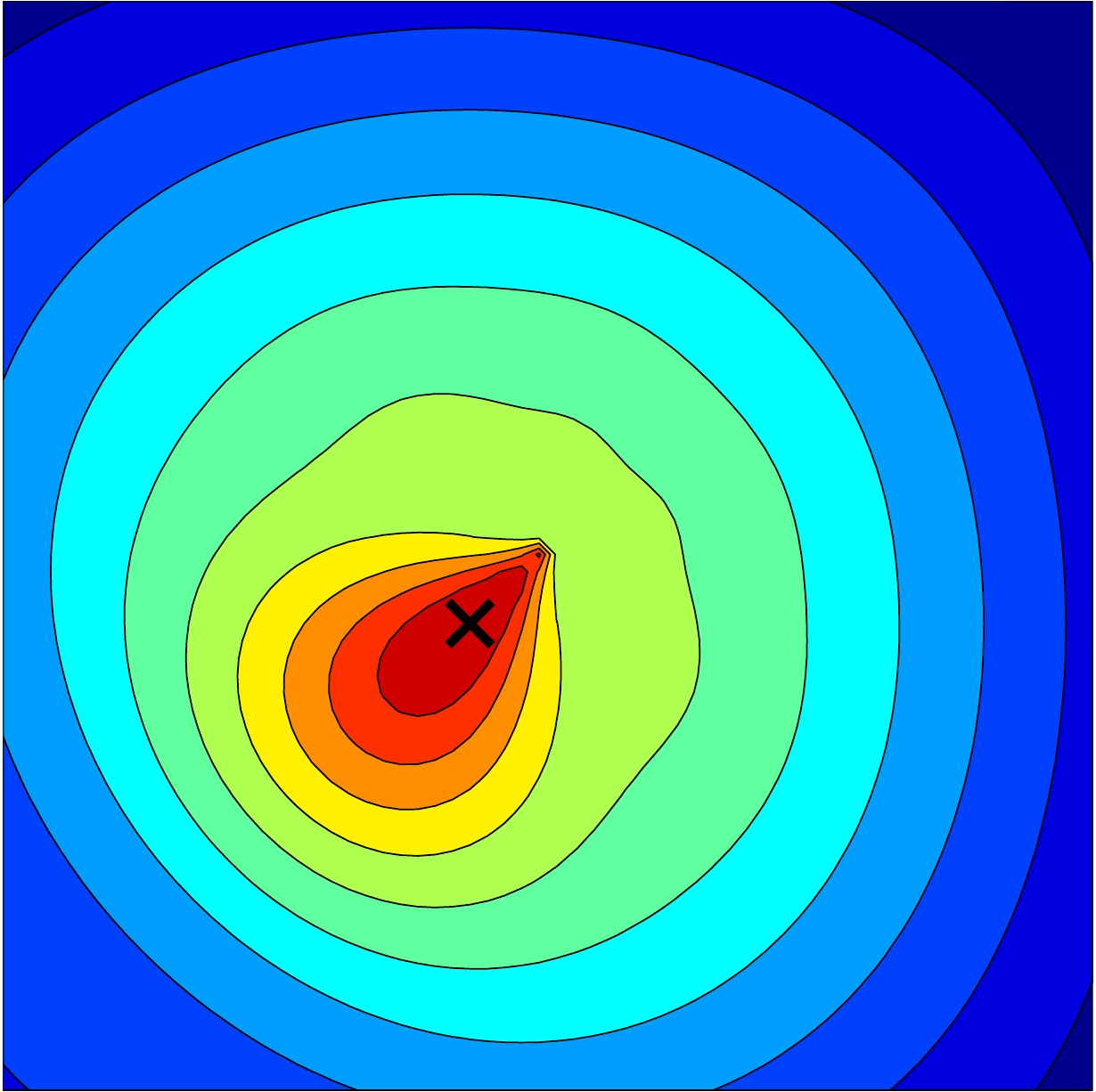}&
   \includegraphics[height=58pt]{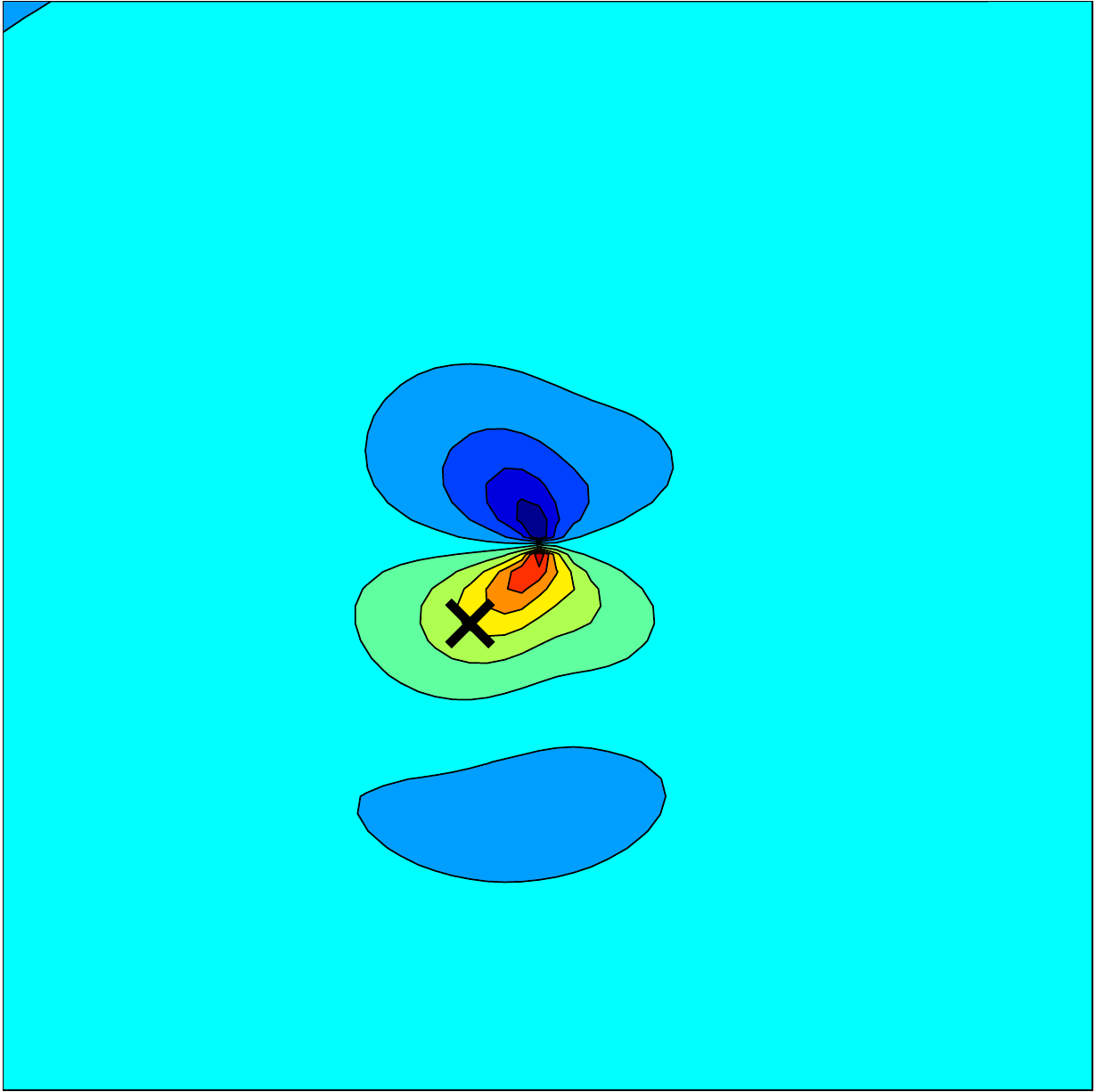}&
   \includegraphics[height=58pt]{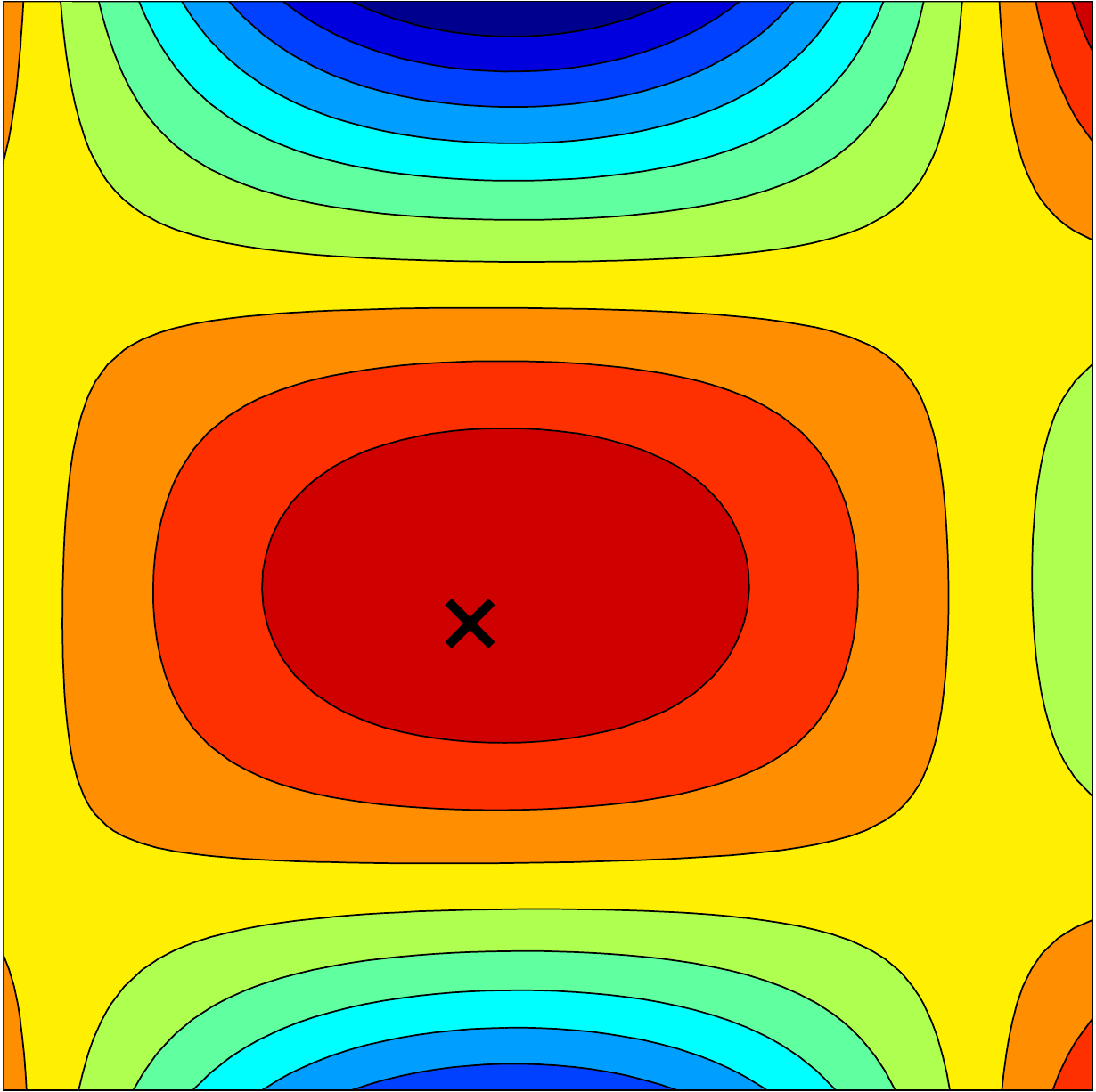}&
   \includegraphics[height=58pt]{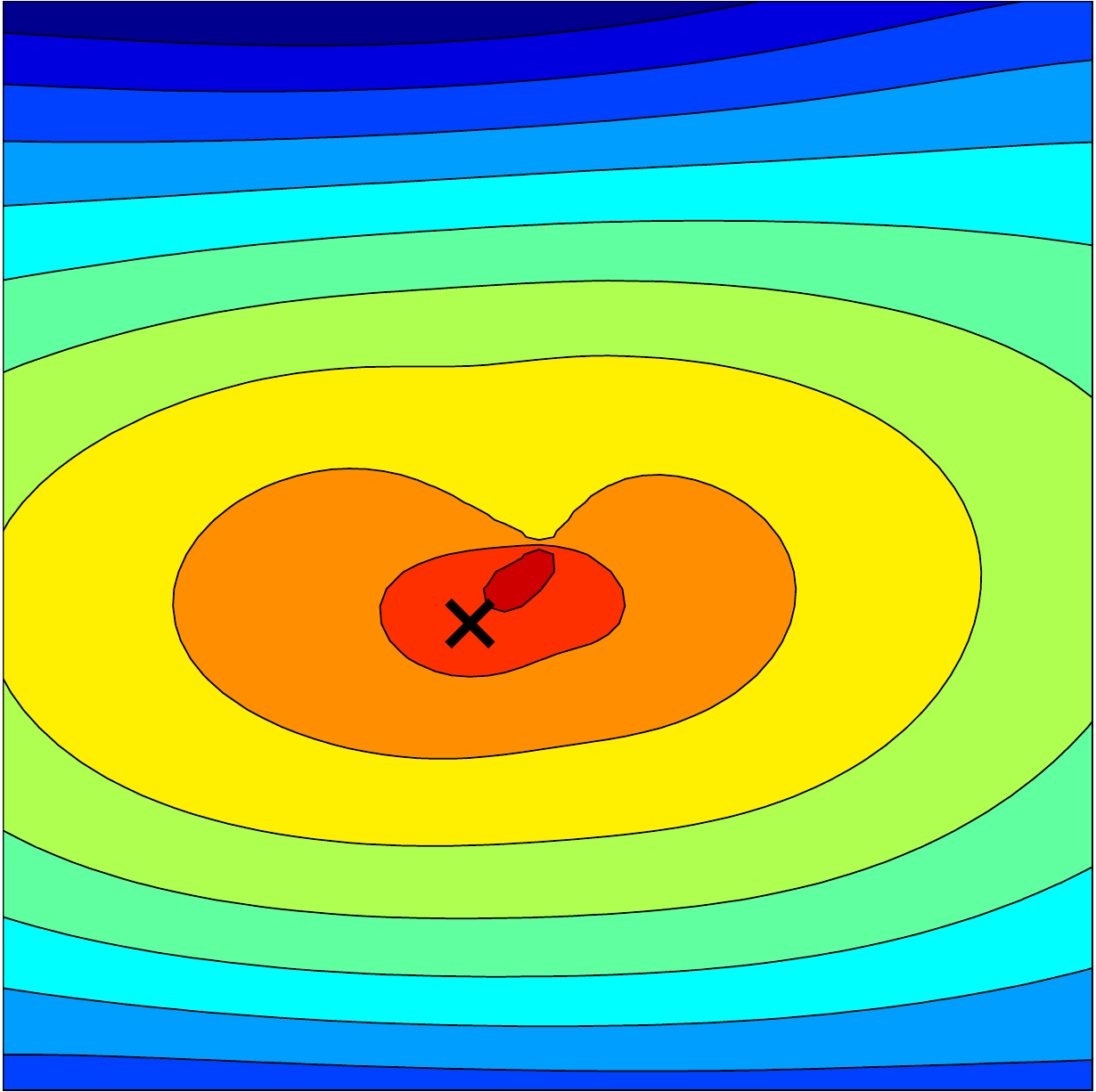} \vspace{8pt}\\
   \kp\hspace{-2pt}\kr\hspace{-2pt}\ktt  & \kx\hspace{-2pt}\ky\hspace{-2pt}\kt  & \kp\hspace{-2pt}\kr\hspace{-2pt}\ktt\hspace{-2pt}+\hspace{-0pt}\kx\hspace{-2pt}\ky\hspace{-2pt}\kt & \kp\hspace{-2pt}\kr\hspace{-2pt}\ktt (LW)& \kx\hspace{-2pt}\ky\hspace{-2pt}\kt (LW) & \kp\hspace{-2pt}\kr\hspace{-2pt}\ktt\hspace{-2pt}+\hspace{-0pt}\kx\hspace{-2pt}\ky\hspace{-2pt}\kt (LW)\\
\end{tabular}
\vspace{15pt}
\caption{Patch maps for different parametrizations and kernels. We present polar and cartesian parametrization separately, and their combination by descriptor concatenation. We present the case for 3 different pixels \p (one pixel per row) shown with ``$\times$''. $\Delta\theta=0$  in all examples, in particular \pt = 0 and \qt=0. The cases without descriptor transformation and with transformation by supervised whitening (LW) are shown.\protect\footref{footfig}
\label{fig:joint}}
\end{figure*}

\vspace{5pt}
In Figure~\ref{fig:joint} we show patch maps for the individual and combined representation, before and after applying learned whitening.
Observe how the combined one better behaves around the center but also how the final similarity is formed after the whitening.

\setcounter{footnote}{3}
\footnotetext{Ten isocontours are sampled uniformly. The similarity is shown in a relative manner and, therefore, the absolute scale is missing (\eg in Figure~\ref{fig:parametrizations} the maximum value is larger in top row compared to bottom due to $\kt(0) > \kt(\nicefrac{\pi}{8})$).\label{footfig}}

\section{Experiments}
\label{sec:exp}
\vspace{5pt}

We evaluate the method on two benchmarks, namely the widely used \emph{Phototourism} (PT) dataset~\cite{WB07}, and the recently released \emph{HPatches} (HP) dataset~\cite{BLVM17}. We first compare the proposed method with the baseline method of Bursuc \etal~\cite{BTJ15} and then with the state-of-the-art methods on the two datasets.
In all our experiments with descriptor post-processing the dimensionality is reduced to 128 except for the cases where the input descriptor is already of lower dimension.

\vspace{10pt}
\textbf{Datasets and protocols.} The Phototourism dataset contains three sets of patches, namely, Liberty (Li), Notredame (No) and Yosemite (Yo). Additionally, labels are provided to indicate the 3D point that the patch corresponds to, thereby providing supervision.
It has been widely used for training and evaluating local descriptors.
Performance is measured by the false positive rate at 95\% of recall (FPR95).
The protocol is to train on one of the three sets and test on the other two. An average over all six combinations is reported.

The HPatches dataset contains local patches of higher diversity, is more realistic, and during evaluation the performance is measured on three tasks: \emph{verification}, \emph{retrieval}, and \emph{matching}.
We follow the standard evaluation protocol~\cite{BLVM17} and report mean Average Precision (mAP).
When evaluating on HP, we follow the protocol and learn the whitening on  PhotoTourism Liberty, or on a pre-defined split of test and train of the HPatches dataset provided by the authors.

\begin{table}[t]
\small
\centering
\setlength\extrarowheight{0pt}
\begin{tabular}{LRRRRRRRR}
\toprule
Test & & & \multicolumn{2}{c}{Liberty} & \multicolumn{2}{c}{Notredame} & \multicolumn{2}{c}{Yosemite} \\ \cmidrule(lr){4-5}\cmidrule(lr){6-7}\cmidrule(lr){8-9}
Train & D & Mean & No& Yo& Li & Yo & No & Li  \\\midrule
\baseline \text{~\cite{BTJ15}}          & 175 &      22.42   &      24.34   &      24.34   &      16.06   &      16.06   &      26.85   &      26.85   \\
\xya                                    & 63  &      35.87   &      34.06   &      34.06   &      34.10   &      34.10   &      39.47   &      39.47   \\
\baseline + \xya                        & 238 &      25.37   &      26.16   &      26.16   &      20.04   &      20.04   &      29.91   &      29.91   \\
\baseline+PCA \text{~\cite{BTJ15}}      & 128 &       8.30   &      12.09   &      13.13   &       5.16   &       5.41   &       7.52   &       6.49   \\
\baseline \text{~\cite{BTJ15}}+LW       & 128 &       7.06   &       8.55   &      10.48   &       4.40   &       3.94   &       8.86   &       6.12   \\
\xya + LW                               & 63  &      15.13   &      17.31   &      20.34   &      10.90   &      11.85   &      16.84   &      13.55   \\
\baseline + \xya + LW                   & 128 &   \bf{5.98}  &   \bf{7.44}  &   \bf{9.84}  &   \bf{3.48}  &   \bf{3.54}  &   \bf{6.56}  &   \bf{5.02}  \\
\midrule[\heavyrulewidth]
\bottomrule

\end{tabular}
\vspace{1.5em}
\caption{Performance comparison on Phototourism dataset between the baseline approach and our combined descriptor. We further show the benefit of learned whitening (LW) over the standard PCA followed by square-rooting. FPR95 is reported for all methods.}\label{tab:ptscores}
\end{table}

\begin{table}[t]
\small
\centering
\setlength\extrarowheight{0pt}
\begin{tabular}{LCCC}
\toprule

\multicolumn{1}{l}{Method} &
\multicolumn{1}{c}{Verification} &
\multicolumn{1}{c}{Matching} &
\multicolumn{1}{c}{Retrieval} \\
\midrule

\baseline \text{~\cite{BTJ15}}           &     80.77   &     32.51   &     48.04   \\
\xya                                     &     70.67   &     15.79   &     30.73   \\
\baseline + \xya                         &     77.97   &     29.34   &     44.23   \\
\baseline+PCA \text{~\cite{BTJ15}}       &     87.11   &     38.45   &     54.81   \\
\baseline \text{~\cite{BTJ15}}+LW        &     88.00   &     41.91   &     58.80   \\
\xya + LW                                &     85.13   &     33.77   &     52.94   \\
\baseline + \xya + LW                    & \bf{88.64}  & \bf{43.81}  & \bf{61.21}  \\

\midrule[\heavyrulewidth]
\bottomrule
\end{tabular}
\vspace{1.5em}
\caption{Performance comparison of the baseline approach and our combined descriptor via mAP on HPatches dataset. PCA and LW are learned on a subset of HP.}\label{tab:hpscores}
\end{table}

\textbf{Comparison with the baseline.} The results of the experimental evaluation are shown in Tables~\ref{tab:ptscores} and \ref{tab:hpscores} for the PT and HP datasets, respectively. For all compared methods, including the baseline, we observed that in the descriptor post-processing stage, the discriminative whitening (marked LW) outperforms PCA followed by square-rooting (originally proposed in~\cite{BTJ15}). The difference is observed among $4^{\text{th}}$ and $5^{\text{th}}$ row of Tables~\ref{tab:ptscores} and \ref{tab:hpscores}.

Polar parametrization with the relative gradient direction (\baseline) significantly outperforms the Cartesian parametrization with the absolute gradient direction (\xya). After the descriptor post-processing (\baseline + LW vs.\ \xya + LW), the gap is reduced. The performance of the combined descriptor (\baseline + \xya) without descriptor post-processing is worse than the baseline descriptor. That is caused by the fact, that the two descriptors are combined with an equal weight, which is clearly suboptimal. No attempt is made to estimate the mixing parameter explicitly, as this is implicitly done in the post-processing stage. The jointly whitened combination of the two parametrizations (last row of Tables~\ref{tab:ptscores} and \ref{tab:hpscores}) consistently outperforms the baseline method.

\textbf{Comparison with the State of the Art.} We compare the performance of proposed method with previously published results on Phototourism dataset in Table~\ref{tab:ptbestscores}. Our method obtains the best performance, while this is achieved with the supervised whitening which is much faster to learn than CNN descriptors. It only takes less than 10 seconds to compute on a modern computer(4 cores, 2.6Ghz) for the \baseline + \xya case on the Phototourism Liberty dataset, as opposed to several hours and GPUs for the deep learning approaches.

The comparison on the HPatches dataset is reported in Table~\ref{tab:hpbestscores}. On the left all methods are considered, independently whether the splits of HPatches have been used for training or not. The table on the right compares only those methods that have {\bf not} used any part of HPatches for training. In this case, the post-processing (LW) of our method was learned on Phototourism Liberty, as done in~\cite{BLVM17} so that the numbers are directly comparable. Note that the proposed method trained on Phototourism Liberty scores high even among the methods that used the split of HPatches in training.

\begin{table}
\small
\centering
\setlength\extrarowheight{0pt}
\def\dd{\hspace{-2pt}-\hspace{-2pt}}
\begin{tabular}{LRRRRRRRR}
\toprule
Test & & & \multicolumn{2}{c}{Liberty} & \multicolumn{2}{c}{Notredame} & \multicolumn{2}{c}{Yosemite} \\ \cmidrule(lr){4-5}\cmidrule(lr){6-7}\cmidrule(lr){8-9}
Train & D & Mean & No& Yo& Li & Yo & No & Li  \\\midrule
DC\dd S2S \text{~\cite{ZK15}}      &   512   &     9.67  &     8.79   &      12.84  &     4.54  &     5.58  &      13.02  &      13.24  \\
DDESC \text{~\cite{STFK+15}}     &   128   &     9.85  &     8.82   &   {\bf8.82} &     4.54  &     4.54  &      16.19  &      16.19  \\
Matchnet \text{~\cite{HLJS+15}}  &   4096  &     7.75  &     6.90   &      10.77  &     3.87  &     5.76  &       8.39  &      10.88  \\
TF\dd M \text{~\cite{BRPM16}}      &   128   &     6.47  & {\bf7.22}  &       9.79  & {\bf3.12} &     3.85  &       7.08  &       7.82  \\
\baseline + \xya + LW            &   128   & {\bf5.98} &     7.44   &       9.84  &     3.48  & {\bf3.54} &   {\bf5.02} &   {\bf6.56} \\

\midrule[\heavyrulewidth]
\bottomrule

\end{tabular}
\vspace{1em}
\caption{Performance comparison with the state of the art on Phototourism dataset. FPR95 is reported for all methods and the best score per dataset is shown in bold.}\label{tab:ptbestscores}
\end{table}

\begin{table}
\centering
\setlength\extrarowheight{0pt}
\def\dd{\hspace{-2pt}-\hspace{-2pt}}
\footnotesize
   \begin{tabular}{@{\sssp}L@{\msp}C@{\bsp}L@{\msp}C@{\bsp}L@{\msp}C@{\sssp}}
      \toprule
      \multicolumn{2}{c}{Verification}   &
      \multicolumn{2}{c}{Matching}       &
      \multicolumn{2}{c}{Retrieval}      \\
      \midrule
      TF\dd R           &      81.92   &    {\bf PCW}   &  {\bf33.69}  &  +TF\dd R       &      40.23  \\
      +TF\dd M          &      82.69   &    +TF\dd M    &      34.29   &  +SIFT          &      40.36  \\
      {\bf PCW}         &  {\bf82.94}  &    +TF\dd R    &      34.37   &  +RSIFT         &      43.84  \\
      +DC\dd S2S        &      83.03   &    +DDESC      &      35.44   &  +DDESC         &      44.55  \\
      +TF\dd R          &      83.24   &    +RSIFT      &      36.77   &  {\bf PCW}      &  {\bf48.26} \\
      {\bf PCW\star}    &  {\bf88.64}  & {\bf PCW\star} &  {\bf43.81}  &  {\bf PCW\star} &  {\bf61.21} \\
     \midrule[\heavyrulewidth]
     \bottomrule
   \end{tabular}
~~~
  \begin{tabular}{@{\sssp}L@{\msp}C@{\bsp}L@{\msp}C@{\bsp}L@{\msp}C@{\sssp}}
      \toprule
      \multicolumn{2}{c}{Verification}   &
      \multicolumn{2}{c}{Matching}       &
      \multicolumn{2}{c}{Retrieval}      \\
      \midrule
      DC\dd S    &      70.04   &  RSIFT      &      27.22   &  DC\dd S2S  &      34.76 \\
      DC\dd S2S  &      78.23   &  DC\dd S2S  &      27.69   &  DC\dd S    &      34.84 \\
      DDESC      &      79.51   &  DDESC      &      28.05   &  TF\dd R    &      37.69 \\
      TF\dd M    &      81.90   &  TF\dd R    &      30.61   &  TF\dd M    &      39.40 \\
      TF\dd R    &      81.92   &  TF\dd M    &      32.64   &  DDESC      &      39.83 \\
      {\bf PCW}  &  {\bf82.94}  &  {\bf PCW}  &  {\bf33.69}  &  {\bf PCW}  &  {\bf48.26}\\
     \midrule[\heavyrulewidth]
     \bottomrule
   \end{tabular}
   \vspace{1em}
   \captionof{table}{Best performing methods on HP dataset. On the left we compare all methods, while on the right only methods
    that have {\bf not} used any part of HPatches for training. The ``+'' refers to ZCA used in~\cite{BLVM17}.
   Our method is noted by PCW (\baseline + \xya + LW ) and shown in bold, while training whitening on a subset of HP is denoted by $\star$. Otherwise it is trained on Liberty (PT). Previously top performing methods are DC-S2S~\cite{ZK15}, DDESC~\cite{STFK+15}, TF~\cite{BRPM16}, and
   RSIFT~\cite{AZ12}. Top 6 methods per task are ranked and shown. Full list of methods in~\cite{BLVM17}.\label{tab:hpbestscores}}
\end{table}

\section{Conclusions}
We have proposed a multiple-kernel local-patch descriptor combining two parametrizations of gradient position and direction. Each parametrization provides robustness to a different type of patch miss-registration: polar parametrization for noise in the dominant orientation, Cartesian for imprecise location of the feature point. Learning a discriminative whitening implicitly sets the relative weight between the two representations. The proposed method consistently outperforms prior methods on two datasets and three tasks.

\vspace{10pt}

\noindent \textbf{Acknowledgments} \hspace{1pt} The authors were supported by the MSMT LL1303 ERC-CZ grant, Arun Mukundan was supported by the CTU student grant SGS17/185/OHK3/3T/13.

\clearpage\newpage
\small
\bibliography{egbib}
\end{document}